\documentclass[final]{cvpr}
\usepackage{times}
\usepackage{epsfig}
\usepackage{graphicx}
\usepackage{amsmath}
\usepackage{amssymb}
\usepackage{mathtools}
\usepackage{float}
\usepackage{enumitem}
\usepackage{widetext}
\usepackage[resetlabels]{multibib}

\newcites{main}{References}
\newcites{sm}{Supplemental References}

\usepackage[strict]{changepage}

\usepackage{framed}

\usepackage[pagebackref=true,breaklinks=true,colorlinks,bookmarks=false]{hyperref}

\def\cvprPaperID{10506} 
\def\confYear{CVPR 2021}

\begin{document}

\sloppy

\DeclarePairedDelimiter\ceil{\lceil}{\rceil}
\DeclarePairedDelimiter\floor{\lfloor}{\rfloor}

\newcommand{\norm}[1]{\left\lVert#1\right\rVert}
\definecolor{antiquefuchsia}{rgb}{0.57, 0.36, 0.51}
\newcommand{\pink}[1]{{\textcolor{antiquefuchsia}{#1}}}

\newcommand{\focusnote}[1]{{\textcolor{purple}{CHECK: #1}}}
\newcommand{\ldnote}[1]{{\textcolor{red}{LD: #1}}}
\newcommand{\mbnote}[1]{{\textcolor{blue}{MB: #1}}}

\definecolor{formalshade}{rgb}{0.95,0.95,0.95}
\definecolor{formalshaderule}{rgb}{0.4,0.4,0.4}

\newenvironment{formal}{%
  \def\FrameCommand{%
    \hspace{1pt}%
    {\color{formalshaderule}\vrule width 2pt}%
    {\color{formalshade}\vrule width 4pt}%
    \colorbox{formalshade}%
  }%
  \MakeFramed{\advance\hsize-\width\FrameRestore}%
  \noindent\hspace{-4.55pt}
  \begin{adjustwidth}{}{7pt}%
  \vspace{2pt}\vspace{2pt}%
}
{%
  \vspace{2pt}\end{adjustwidth}\endMakeFramed%
}

\title{CellSegmenter: unsupervised representation learning and \\instance segmentation of modular images}

\author{Luca D'Alessio\\
  Data Sciences Platform, Broad Institute\\
  415 Main St, Cambridge, MA 02142 \\
  {\tt\small ldalessi@broadinstitute.org}
  \and
  Mehrtash Babadi\\
  Data Sciences Platform, Broad Institute\\
  415 Main St, Cambridge, MA 02142 \\
  {\tt\small mehrtash@broadinstitute.org}
}

\maketitle



\begin{abstract}
We introduce CellSegmenter, a structured deep generative model and an amortized inference framework for unsupervised representation learning and instance segmentation tasks. The proposed inference algorithm is convolutional and parallelized, without any recurrent mechanisms, and is able to resolve object-object occlusion while simultaneously treating distant non-occluding objects independently. This leads to extremely fast training times while allowing extrapolation to arbitrary number of instances. We further introduce a transparent posterior regularization strategy that encourages scene reconstructions with fewest localized objects and a low-complexity background. We evaluate our method on a challenging synthetic multi-MNIST dataset with a structured background and achieve nearly perfect accuracy with only a few hundred training epochs. Finally, we show segmentation results obtained for a cell nuclei imaging dataset, demonstrating the ability of our method to provide high-quality segmentations while also handling realistic use cases involving large number of instances.
\end{abstract}

\section{Introduction}
Object recognition and localization is the essence of scene understanding, a highly complex feat of human intelligence. We rely on basic notions of physics such as conservation, continuity, and causality, as well as repeated and similar experiences to tackle this problem. Formalizing the minimal structure required to perform this task is the subject of unsupervised instance segmentation and has a long history of active research and innovation.

Significant headway has been made recently using {\em structured} and {\em deep} generative models~\citemain{AIR,genesis,greff2019multi,monet}: similar to classical Bayesian networks, the structure acts as an inductive bias, enforcing basic notions of physics, geometry, and scene composition, whereas the deep neural components take on the heavy lifting role of object recognition and pattern generation, within the structural confines of the model. Inference and learning are typically done recurrently, as a sequence of detect-crop-encode-decode-paste operations. While being an intuitively appealing paradigm, recurrent recognition leads to major practical drawbacks: it is (\textbf{P1}) computationally slow and scales poorly with the number of objects, while also suffering from (\textbf{P2}) poor generalization to unseen number of instances at test time. Another fundamental challenge is that any structured model for scene generation is very likely to be {\em misspecified}, at least for {\em some} data distributions. The notion of ``object'' and ``background'' can assume arbitrarily complex variations, and the evidence for the correct segmentation may very well lie outside of the the domain of still images. In practice, (\textbf{P3}) subtle model misspecification or non-identifiability can override the prior structure and lead to poor inferences; and this issue is only exacerbated by the bias inherent to variational inference. 

In this paper, we propose a {\em convolutional} amortized inference method, free from recurrence, that solves problems (\textbf{P1}) and (\textbf{P2}) while maintaining the desirable aspects of recurrent recognition models, e.g. the ability to resolve object-object occlusion. Furthermore, we propose using posterior regularization to mitigate problem (\textbf{P3}), and as an effective one-shot alternative to expensive hyperparameter tuning. Our developments are inspired by the problem of segmenting cell microscopy images, and we call our method ``CellSegmenter''. Cell segmentation is an active area of applied ML research; it is a problem that suffers from paucity of high-quality labeled data and benefits from novel unsupervised and weakly-supervised developments.\\

\begin{figure*}[h!]
\centering
\includegraphics[width=0.85\textwidth]{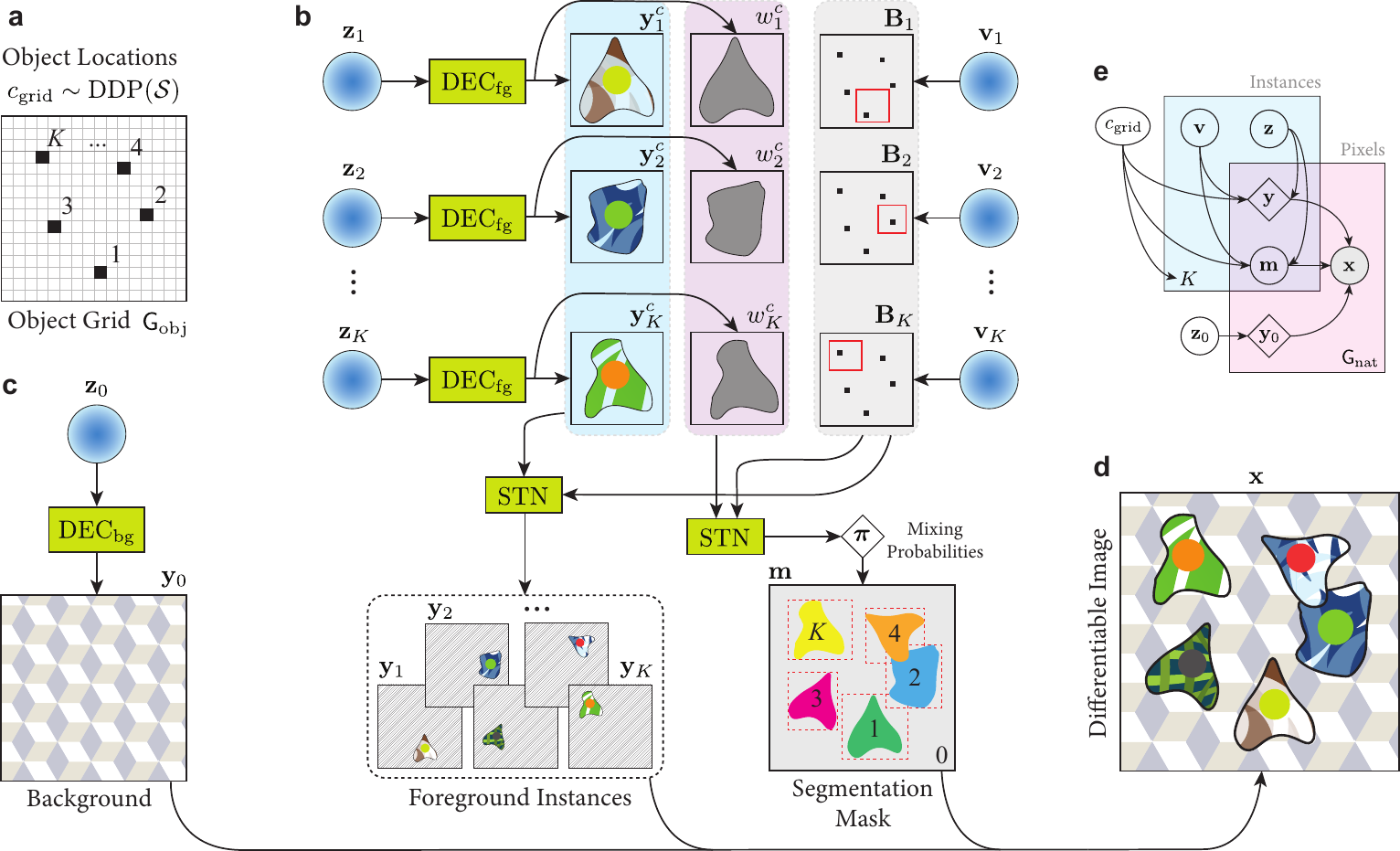}
\caption{\textbf{CellSegmenter generative model.} A probabilistic process for generating differentiable modular images. (a) sampling crude object locations; (b) sampling foreground instances; (c) sampling the background image; (d) image composition; (e) conditional independence structure of the model. Please refer to Sec.~\ref{sec:generative_model} for further details.} 
\label{fig:generative_model}
\end{figure*}

\noindent{\bf Related works---} The related literature is extensive, ranging from classical Bayesian techniques to domain-specific models enriched with various degrees of weak supervision, e.g. with incomplete annotations, temporal data, transfer learning, and domain adaptation. Here, we recount a number of immediately related works inspiring the present paper. In particular, we focus on ``zero-knowledge'' models in which the segmentation is entirely guided by the inductive biases of the structural assumptions and information theoretic considerations. \textit{Attend, Infer, Repeat (AIR)}~\citemain{AIR} treats inference as an iterative process. A recurrent neural network (RNN) sequentially makes region proposals which are further processed by an ensemble of independent variational auto-encoders~\citemain{VAE_kingma2013} (VAEs). {\em AIR} establishes the possibility of successful scene understanding using structured deep generative models. \textit{SuPAIR}~\citemain{faster-AIR} is a variant of {\em AIR} which uses sum-product networks~\citemain{spn1, spn2} (SPNs) instead of VAEs, resulting in faster operation. In addition, {\em SuPAIR} explicitly models the background component, the absence of which is a caveat of \textit{AIR}. \textit{GENESIS}~\citemain{genesis}, \textit{IODINE}~\citemain{greff2019multi} and \textit{MONet}~\citemain{monet} introduce spatial attention masks. In a recurrent setup, each attention mask proposes a previously unexplained region of the scene and presents it to a VAE. The information bottleneck imposed by the VAE forms the basis of representation learning and recognition. These intriguing approaches allow generation of complex scenes. However, the obtained segmentations are rather arbitrary and often associate several localized objects to the same instance. Moreover, recurrent models are fundamentally hampered by (\textbf{P1}) and (\textbf{P2}). There has been tangible progress in replacing sequential attention with global attention, e.g. as in \textit{MIST}~\citemain{mist_2018} and \textit{MVAE}~\citemain{nash17}. The generators resemble the previous models, however, object proposals are obtained by selecting top-$K$ voxels of an auxiliary attention map, with $K$ being treated as a fixed parameter. As such, neither model ``learns to count''. {\em MVAE} proposes a cross validation algorithm to estimate the number of instances by counting the number of attention map peaks exceeding a threshold. Finally, we are unaware of works toward mitigating model misspecification in the context of unsupervised scene understanding (\textbf{P3}).

\noindent{\bf Novel contributions ---} We briefly recount the main contributions of our work here for reference, along with pointers to the main text:

\textbf{(1)} In Sec.~\ref{sec:generative_model}, we propose a structured deep differentiable image generator that captures the basic phenomenology of a large class of modular images: an unknown number of localized instances on the backdrop of a structured background, both having a low descriptive complexity. Prior preference is given to configurations with non- or weakly- occluding objects using a spatial determinantal point process (DPP)~\citemain{kulesza2012determinantal_main}.\\
\indent\textbf{(2)} In Sec.~\ref{sec:inference}, we propose a fully convolutional and parallelized inference framework by {\em repurposing} the U-Net architecture~\citemain{unet_original_main}, traditionally used for supervised semantic segmentation, as a {\em variational posterior amortizer}. The proposed inference framework is fast and scalable and is able to resolve object-object interactions.\\
\indent\textbf{(3)} In Sec.~\ref{sec:learning}, we propose a simplified and asymmetric variant of the recently introduced Generalized ELBO Constrained Optimization (GECO) method~\citemain{GECO} for imposing arbitrary inequality constraints on the posterior space as a drop-in replacement for the commonly used ELBO loss. We find that imposing simple posterior constraints, such as lower and upper bound on the size and density of foreground instances, can very effectively mitigate model misspecification, obviating costly hyperparameter tuning and contrived training schedules.\\
\indent\textbf{(4)} In Sec.~\ref{sec:graph_segmentation}, we introduce a principled post-processing method, based on graph community detection, for combining a collection of probabilistic {\em instance} segmentations (e.g. posterior samples, overlapping sliding windows) and obtaining a global {\em consensus} instance segmentation. This technique allows us to process very large contiguous images, such as cell microscopy and aerial images, in small sliding windows and obtain a contiguous segmentation in a rapid map-reduce framework, without resorting to heuristic merging operations.\\

\noindent{\bf Preliminaries and notation ---} We briefly introduce key notations here. An image with height $H$, width $W$, and $C$ channels is denoted by $\mathbf{x} \in \mathsf{IM}$ where $\mathsf{IM} \equiv \mathbb{R}^{H\times W \times C}$. The background and the foreground images are denoted by $\mathbf{y}_0\in \mathsf{IM}$ and $\mathbf{y}_{1:K}\in \mathsf{IM}$, respectively. Each pixel has a mixture probability of belonging to each component, denoted by $\boldsymbol{\pi}_{0:K}$. We refer to the Cartesian grid of points at the native image resolution as $\mathsf{G}_\mathrm{nat}$, and further define a coarser grid $\mathsf{G}_\mathrm{obj}$ with spacing that is roughly determined by the expected size of the smallest object $\sim \ell^<_\mathrm{obj}$ (see Fig.~\ref{fig:generative_model}{\bf a}). $\mathsf{G}_\mathrm{obj}$ contains $\tilde{H} \times \tilde{W}$ points, where $\tilde{H} = \ceil{H/\ell^<_\mathrm{obj}}$ and $\tilde{W} = \ceil{W/\ell^<_\mathrm{obj}}$.

\section{Generative Model}\label{sec:generative_model}
Our proposed generative model is illustrated in Fig.~\ref{fig:generative_model} and comprises four main stages:\\

\noindent{\bf Crude object locations ---} The crude spatial locations of all objects are sampled simultaneously over the coarse grid $\mathsf{G}_\mathrm{obj}$ via a Determinantal Point Process (DPP)~\citemain{kulesza2012determinantal_main} with an RBF kernel $\mathcal{S}^{l,m} = \rho\,\exp\big[-\norm{\mathbf{r}_l - \mathbf{r}_m}_2^2 / (2\,{\ell}^2)\big]$ where $\mathbf{r}_l, \mathbf{r}_m \in \mathsf{G}_\mathrm{obj}$. The DPP is a repulsive point process, induces negative spatial correlation, and leads to a scene containing few object-object occlusions. More explicitly, we define the $c_\mathrm{grid}: \mathsf{G}_\mathrm{obj} \rightarrow \{0, 1\}$ as a binary random field over $\mathsf{G}_\mathrm{obj}$ with a DPP prior, see Fig.~\ref{fig:generative_model}{\bf a}:
\begin{equation}
    c_\mathrm{grid} \sim \mathrm{DPP}(\mathcal{S}).
\end{equation}
A brief overview of DPP is provided in Suppl. Mat \S1. Note that the number of instances $K \equiv \sum_{\mathbf{r} \in \mathsf{G}_\mathrm{obj}} c_\mathrm{grid}(\mathbf{r})$ is a stochastic variable, the statistics of which is controlled by the parameters of the RBF kernel.\\

\noindent{\bf Foreground bounding boxes, appearances, and local mixing weights ---} We define each foreground object as a bundle of {\em bounding box}, {\em appearance}, and {\em local mixing weights}.  To obtain the bounding boxes, we sample an {\em instance location latent codes} $\mathbf{v}_{1:K} \sim \mathcal{N}(\mathbf{0}_4, \mathbf{I}_{4\times4})$. These variables encode the size of the bounding boxes, their fine scale placement, and together with the crude object locations provided by $c_\mathrm{grid}$, define the foreground object bounding boxes (see Suppl. Mat. \S2 for details): 
\begin{equation}
\mathbf{B}_k = \left(b^\mathrm{x}_k, b^\mathrm{y}_k, b^\mathrm{w}_k, b^\mathrm{h}_k\right),\quad k=1,\dots,K
\end{equation}
At this point, we have a collection of $K$ weakly overlapping bounding boxes. To proceed, we sample the {\em instance appearance latent codes} $\mathbf{z}_{1:K} \sim \mathcal{N}(\mathbf{0}_{D_\mathrm{fg}}, \mathbf{1}_{D_\mathrm{fg} \times D_\mathrm{fg}})$ and transform them to obtain (1) a rendering of the instance appearance, $\mathbf{y}^c_{1:K} \in \mathsf{IM}^c \equiv \mathbb{R}^{H_\mathrm{fg} \times W_\mathrm{fg} \times C}$, and (2) mixing weights $\mathbf{w}^c_k \in \mathsf{W}^c \equiv [0, 1)^{H_\mathrm{fg} \times W_\mathrm{fg}}$:
\begin{equation}\label{eq:dec_fg}
    \mathbf{y}^c_k, \mathbf{w}^c_k = \mathrm{DEC}_\mathrm{fg}(\mathbf{z}_k).
\end{equation}
The mixing weights can be understood as soft segmentation masks for each instance. All foreground objects invoke the same $\mathrm{DEC}_\mathrm{fg}$ decoder (see Suppl. Mat. \S3). The generative process of foreground instances is shown in Fig.~\ref{fig:generative_model}{\bf b}.\\

\noindent{\bf Background image ---} We assume that the background image is independent of the foreground objects\footnote{This assumption can be easily relaxed by conditioning the latent code of the background on the concatenation of the latent code of foreground instances. The relevance of this assumption depends on one's notion of background, e.g. if it is desired to consider the ``shadow'' of objects as part of the background and not the appearance of the objects, or vice versa.}. We obtain the background image by sampling a background latent code $\mathbf{z}_0 \sim \mathcal{N}(\mathbf{0}_{D_\mathrm{bg}}, \mathbf{1}_{D_\mathrm{bg} \times D_\mathrm{bg}})$ and transforming it to the image space via a differentiable map $\mathrm{DEC}_\mathrm{bg}: \mathbb{R}^{D_\mathrm{bg}} \rightarrow \mathsf{IM}$:
\begin{equation}\label{eq:dec_bg}
    \mathbf{y}_0 = \mathrm{DEC}_\mathrm{bg}(\mathbf{z}_0).
\end{equation}
We implement $\mathrm{DEC}_\mathrm{bg}$ as a simple expansive feed-forward CNN; see Suppl. Mat. \S4 for details.\\

\noindent{\bf Scene composition ---} A Spatial Transformer Network (STN)~\citemain{STN} embeds, in a differentiable way, the instance appearance and mixing weights into the image space according to the bounding boxes $\mathbf{B}_{1:K}$:
\begin{equation}
\label{eq:mixing_weights}
\begin{split}
\mathbf{y}_{1:K} &= \mathrm{STN}(\mathbf{y}^c_{1:K}\, |\, \mathbf{B}_{1:K}), \\
\mathbf{w}_{1:K} &= \mathrm{STN}(\mathbf{w}^c_{1:K}\, |\, \mathbf{B}_{1:K}).
\end{split}
\end{equation}
Note that $\mathbf{y}_{1:K}$ and $\mathbf{w}_{1:K}$ are identically zero outside of their corresponding bounding boxes. Up to this point, the instances have been treated independently. We now introduce the {\em mixing probabilities} $\boldsymbol{\pi} \in (K\mathrm{-simplex})^{H \times W}$ as follows:
\begin{equation}
    \boldsymbol{\pi}_{1:K} =  \frac{\mathbf{w}_{1:K}}{ \mathrm{max}\left(1\,, \sum_{j=1}^K \mathbf{w}_j \right)}, \qquad \boldsymbol{\pi}_0 = 1 - \sum_{j=1}^K \boldsymbol{\pi}_{j},
    \label{eq:weights_to_masks}
\end{equation}
where the arithmetic is understood in pixel-wise sense. The proposed mapping from mixing weight $\mathbf{w}_{1:K}$ to mixing probabilities $\boldsymbol{\pi}_{0:K}$ entails the following favorable properties: (1) the mixing probability for each object is strictly zero outside of its bounding box; (2) for all pixels, the sum of the foreground mixing probabilities is strictly smaller than 1, so that the background mixing probability $\boldsymbol{\pi}_0$ could be simply defined as a complement of total foreground probability; (3) for pixels that are covered by a single bounding box, the expression defaults to independent objects; (4) for pixels that are covered by two or more bounding boxes, the global mixing probability becomes a normalized mixture of local probabilities between the involved instances.
We obtain the {\em quantized segmentation mask} for a pixel $(i, j) \in \mathsf{G}_\mathrm{nat}$ via an Categorical sampling from $\boldsymbol{\pi}$:
\begin{equation}
    \mathbf{m}^{(i,j)} \, \big| \, \boldsymbol{\pi}^{(i,j)} \sim \mathrm{Categorical}(\boldsymbol{\pi}^{(i,j)}).
    \label{eq:categorical}
\end{equation}
We note that even though the sampling is performed independently for each pixel, the resulting segmentation mask is highly correlated across nearby pixels due to the correlation structure built into $\boldsymbol{\pi}$. Ultimately, the composed scene is obtained as:
\begin{equation}
    \mathbf{x}^{(i,j)} \, \Big| \, m^{(i,j)}_{0:K}, \mathbf{y}^{(i,j)}_{0:K} \sim \mathcal{N}\left({\textstyle  \sum_{q=0}^K} m^{(i,j)}_q\,\mathbf{y}^{(i,j)}_{q}, \sigma\right).
\end{equation}
The scene composition stage is schematically shown in Fig.~\ref{fig:generative_model}{\bf d}. The conditional independence structure of the model is shown in panel {\bf e} for reference.



\subsection{Amortized Variational Inference}\label{sec:inference}
\begin{figure*}
\centering
\includegraphics[width=\textwidth]{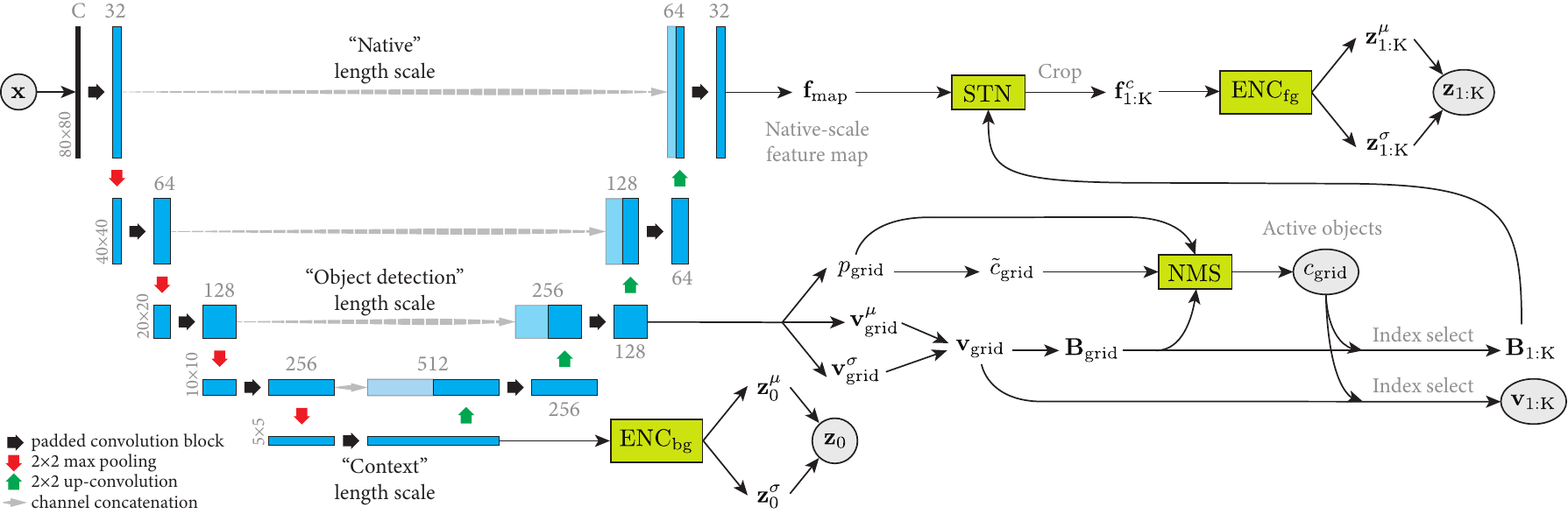}
\caption{\textbf{CellSegmenter amortized variational inference model.} Please refer to Sec.~\ref{sec:inference} for further details.}
\label{fig:inference_model}
\end{figure*}

Devising a variational inference (VI) strategy for the present model is a very challenging problem due to several factors: (1) the model involves several {\em local} latent variables and as such, traditional mean-field VI using factorized Gaussian posteriors~\citemain{VI_factorized, VI_factorized_jaakkola200110} is impractically space- and time-inefficient; crucially, the inference of all local latent variables (e.g. $\mathbf{v}_{1:K}$, $\mathbf{z}_{1:K}$, etc) must be {\em amortized} in the spirit of VAEs~\citemain{VAE_kingma2013, vae-deep}; (2) the cardinality of local latent variables (e.g. number of foreground instances) varies from image to image; (3) the binary random field $c_\mathrm{grid}$ assumes $2^{|\mathsf{G}_\mathrm{obj}|}$ possible configurations for each image, preventing the best practice of variable elimination via full enumeration~\citemain{obermeyer2019tensor}.

Our proposed fully-amortized VI framework is shown in Fig.~\ref{fig:inference_model} and is inspired by the U-Net architecture~\citemain{unet_original_main}. In standard applications, U-Net is used for supervised semantic segmentation and is trained {\em directly} on pairs of training segmentation masks and raw images. Here, in contrast, the U-Net module is trained {\em indirectly}, by extracting features at different length scales, playing the role of the parameter amortizer of a variational posterior ansatz, and with the ultimate goal of increasing the model marginal likelihood.


Features at the bottom of the U-Net encode the most global information; as such, they can amortize the inference of the background latent code $\mathbf{z}_0$ (see Fig.~\ref{fig:inference_model}):
\begin{equation}
    \mathbf{z}_0 \, | \, \mathbf{x} \sim \mathcal{N}(\mathbf{z}_0^\mu[\mathbf{x}], \mathbf{z}_0^\sigma[\mathbf{x}]),
\end{equation}
where $\mathbf{z}_0^{\mu(\sigma)}[\mathbf{x}]$ are obtained from a lightweight encoder network (see Suppl. Mat \S4 for details). Object detection is amortized using intermediate-scale features, corresponding to the resolution of $\mathsf{G}_\mathrm{obj}$. We extract $p_\mathrm{grid} \in [0, 1]^{\tilde{H} \times \tilde{W}}$, $\mathbf{v}_\mathrm{grid}^{\mu} \in \mathbb{R}^{\tilde{H} \times \tilde{W}}$, and $\mathbf{v}_\mathrm{grid}^{\sigma} \in \mathbb{R}_{+}^{\tilde{H} \times \tilde{W}}$ from these features via a MLP followed by appropriate nonlinear activations. These gridded quantities are thought of as object presence probability map, posterior mean, and posterior variance of all bounding box proposals, respectively, and parameterize variational posteriors distributions for $c_\mathrm{grid}$ and $\mathbf{v}_\mathrm{grid}$. More concretely, $\tilde{c}_\mathrm{grid} \, | \, \mathbf{x} \sim \mathrm{Bernoulli}(p_\mathrm{grid}[\mathbf{x}])$, and $\mathbf{v}_\mathrm{grid} \, | \, \mathbf{x} \sim \mathcal{N}\big(\mathbf{v}_\mathrm{grid}^{\mu}[\mathbf{x}], \mathbf{v}_\mathrm{grid}^{\sigma}[\mathbf{x}]\big)$. 
In its present form, the binary random field $\tilde{c}_\mathrm{grid}$ might contain nearby grid points that recognize the same object with high probability multiple times. In the forward model, the negative correlation induced by DPP penalizes such undesirable configurations. To mirror this prior structure, we supplement the variational posterior with a non-max suppression (NMS) operator~\citemain{nms_example} as a mechanism to remove redundant proposals:
\begin{equation}\label{eq:nms}
    c_\mathrm{grid} \, | \, \mathbf{x} \leftarrow \mathrm{NMS}(\tilde{c}_\mathrm{grid} \, | \, p_\mathrm{grid}, \mathbf{B}_\mathrm{grid}[\mathbf{v}_\mathrm{grid}]).
\end{equation}
In brief, the NMS operator acts as a hard filter by calculating the overlap between the bounding boxes corresponding to $\tilde{c}_\mathrm{grid} = 1$ and removing the lower confidence proposals from all the pairs that overlap beyond a specified intersection-over-minimum (IoM) threshold $\alpha$ (see Suppl. Mat. Sec. \S5 for details). The number of proposals passing this filtering procedure determines $K$, and their grid indices are used to select $\mathbf{v}_{1:K} \, | \, \mathbf{x}$ and $\mathbf{B}_{1:K} \, | \, \mathbf{x}$ from $\mathbf{v}_\mathrm{grid} \, | \, \mathbf{x}$ and $\mathbf{B}_\mathrm{grid} \, | \, \mathbf{x}$ respectively, see Fig.~\ref{fig:inference_model}. Equipped with a set of weakly overlapping region proposals, we can take the final step of determining the foreground appearance latent codes. To this end, we crop the feature map
at end of the U-Net (defined at native resolution) according to each of the bounding boxes $\mathbf{B}_{1:K}$ using STN, and obtain the latent encoding through a shared foreground encoder $\mathrm{ENC}_\mathrm{fg}: \mathbb{R}^{H_\mathrm{fg} \times W_\mathrm{fg} \times C_\mathrm{U}} \rightarrow \mathbb{R}^{D_\mathrm{fg}} \times \mathbb{R}_{+}^{D_\mathrm{fg}}$ (where $C_\mathrm{U}$ is the number of terminal U-Net channels; $C_\mathrm{U}=32$ in Fig.~\ref{fig:inference_model}):
\begin{equation}
\begin{split}
    \mathbf{f}^c_{1:K}[\mathbf{x}]& = \mathrm{STN}\left(\mathbf{f}_\mathrm{map}[\mathbf{x}] \, | \, \mathbf{B}_{1:K}\right), \\
    \mathbf{z}^\mu_{1:K}[\mathbf{x}], \mathbf{z}^\sigma_{1:K}[\mathbf{x}] & = \mathrm{ENC}_\mathrm{fg} \left(\mathbf{f}^c_{1:K}[\mathbf{x}]\right), \\
    \mathbf{z}_{1:K} \, | \, \mathbf{x} & \sim \mathcal{N}\left(\mathbf{z}^\mu_{1:K}[\mathbf{x}], \mathbf{z}^\sigma_{1:K}[\mathbf{x}]\right).
\end{split}
\label{eq:crop_f_map}
\end{equation}
As highlighted earlier, all of the instances are treated in parallel and yet, the latent codes of spatially close instances {\em can} be correlated: the feature map at the end of the U-Net carries information from different length scales and in particular, contains information about the recognition probability of all objects and the background; as such, the network can use the feature map to resolve partial occlusions and subtract the background, and ultimately improve the operation of a lightweight foreground encoder $\mathrm{ENC}_\mathrm{fg}$ (see Suppl. Mat. \S6 for details).

We conclude with section with a few quick remarks: {\bf (1)} the utilization of U-Net as a multi-scale feature extractor, and crucially, using the features extracted at different length scales to amortize appropriate latent variables, is a very versatile design choice for designing amortized VI guides for modular images; {\bf (2)} the choice of extracting the object latent codes from the native-resolution feature map at the end of the U-Net, in contrast to directly cropping the input image (e.g. as in {\em AIR}~\citemain{AIR}), is a crucial advantage of our method; {\bf (3)} we emphasize that our inference model contains a NMS operation and yet, it is trained end-to-end via standard gradient-based optimization. The situation is not different from training models having max pooling units: the NMS operation acts as a hard filter that stops certain proposals from being processed while passing through others. The gradients that back-propagate trough the passing proposals are used to train proposal probabilities and all the network weights which are shared among all instances.


\section{Learning\label{sec:learning}}
The canonical learning objective in the stochastic variational inference (SVI) framework is the maximization of the evidence lower bound (ELBO)~\citemain{black_box_VI} over random mini-batches of data via gradient-based methods. The loss is $\mathcal{L}_\mathrm{ELBO} = -\mathbb{E}_{\mathbf{x} \sim p_X(\mathbf{x})} \, \mathbb{E}_{\mathbf{Z} \sim q_\phi(\mathbf{Z};\mathbf{x})}[\log\,p_\theta(\mathbf{x}, \mathbf{Z}) - \log\,q_\phi(\mathbf{Z};\mathbf{x})]$, where $q_\phi(\mathbf{Z};\mathbf{x})$ is the backward (inference) model, $p_\theta(\mathbf{x}, \mathbf{Z})$ is the forward (generative) joint distribution,  $\mathbf{Z}$ is the bundle of latent variables, and $\theta$ and $\phi$ denote all trainable parameters of the generative and inference processes, respectively (see Suppl. Mat. \S7 for a full glossary). The ELBO loss can be conveniently rearranged in terms of reconstruction and Kullback-Leibler (KL) divergence contributions, $\mathcal{L}_\mathrm{ELBO} = \mathcal{L}_\mathrm{rec} + \mathcal{L}_\mathrm{KL}$. The posterior expectation over continuous latent variables is usually approximated with a single reparameterized (i.e. differentiable w.r.t. $\theta$ and $\phi$) Monte Carlo (MC) sample~\citemain{kingma_auto-encoding_2013}, however, special care is required for the discrete latent variables. Here, we perform enumeration over the mask component ${m}_{0:K}$ independently for each pixel and estimate the posterior expectation over $c_\mathrm{grid}$ using a single MC sample, however, endowed with the straight-through gradient estimator (cf. Ref.~\citemain{straight_throug} for a review). 
More concretely, we have:
\begin{eqnarray}\label{eq:loss_explicit}
&&\hspace{-30pt}\textstyle{\mathcal{L}_\mathrm{rec} = \pink{\frac{1}{|\mathsf{G}_\mathrm{nat}|}}\frac{1}{2\,\sigma^2}\sum_{p \in \mathsf{G}_\mathrm{nat}} \sum_{k=0}^{\hat{K}} \hat{\pi}_k^{p}\,\big\lVert\mathbf{x}^{p} - \hat{\mathbf{y}}_k^{p}\big\rVert_2^2},\nonumber\\
&&\hspace{-30pt}\textstyle{\mathcal{L}_\mathrm{KL} = \pink{\frac{1}{D_\mathrm{bg}}}\,f^\mathcal{N}_\mathrm{KL}(\hat{\mathbf{z}}_\mathrm{0}^\mu, \hat{\mathbf{z}}_\mathrm{0}^\sigma) + \pink{\frac{1}{D_\mathrm{fg}\,\hat{K}}}\,\sum_{k=1}^{\hat{K}}f^\mathcal{N}_\mathrm{KL}(\hat{\mathbf{z}}_\mathrm{k}^\mu, \hat{\mathbf{z}}_\mathrm{k}^\sigma)}\nonumber\\
&&\hspace{-30pt}\textstyle{+ \pink{\frac{1}{4\,\hat{K}}}\,\sum_{k=1}^{\hat{K}}f^\mathcal{N}_\mathrm{KL}(\hat{\mathbf{v}}_\mathrm{k}^\mu, \hat{\mathbf{v}}_\mathrm{k}^\sigma) + \pink{\frac{1}{\mathcal{N}_\mathrm{grid}}}\,f_\mathrm{KL}^\mathrm{grid}}(\hat{c}_\mathrm{grid};\hat{p}_\mathrm{grid}).
\end{eqnarray}
Hereafter, variables with hat are either directly differentiable posterior samples, or are derived from such variables through differentiable operations. The highlighted pre-factors condition each term to be $\mathcal{O}(1)$ and render the qualitative behavior of the model independent of image resolution and the size of latent representations. Furthermore, $f^\mathcal{N}_\mathrm{KL}(\boldsymbol{\mu}, \boldsymbol{\sigma}) \equiv \frac{1}{2}\sum_{i=1} \left(\sigma^2_i + \mu^2_i -1 -\log \sigma^2_i \right)$ is the exact KL divergence between $\mathcal{N}(\boldsymbol{\mu}, \boldsymbol{\sigma})$ and $\mathcal{N}(\mathbf{0}, \mathbf{1})$, and $f_\mathrm{KL}^\mathrm{grid} \equiv D_\mathrm{KL}\left[\mathrm{Cat}(\hat{p}_\mathrm{grid})\, ||\, P_\mathrm{DPP}(\mathcal{S})\right]$ is approximated using a single MC sample, see Suppl. Mat \S8 for details. 
Due to the nonlinearity of DPP, the scaling behavior of $f_\mathrm{KL}^\mathrm{grid}$ is non-trivial, and its normalization, denoted by $\mathcal{N}_\mathrm{grid}$, is estimated using an exponentially-weighted moving average.\\

\noindent{\bf Soft asymmetric posterior regularization (SAPR) ---} A theoretical advantage of structured latent variable models is the ability to impose one's prior beliefs and fundamental structural relationships using judicious choice of priors and conditional independence relationships. In practice, though, subtle model misspecification or non-identifiability can overcome the prior structure and lead to poor inferences~\citemain{mispecification_1, mispecification_2}. The standard approach is to perform extensive and costly hyperparameter tuning and cross-validation to satisfy a number of desired posterior criteria.


The normalization constants introduced in Eq.~\eqref{eq:loss_explicit} is a basic strategy to combat model misspecification and/or sub-optimal choice of hyperparameters by explicitly balancing representation complexity vs. reconstruction fidelity. While being quite effective in many cases, we found this simple rescaling to be insufficient to prevent pathological solutions in more challenging scenarios (e.g. images with highly structured backgrounds). A powerful adjunct strategy for safeguarding Bayesian inference is to directly regularize the posterior space~\citemain{posterior_regularization}. Here, for instance, reasonable posterior regularizations (PR) include imposing lower and upper bounds on the 2D area of objects, number of objects, and scene reconstruction error. Recently, Ref.~\citemain{GECO} has introduced a practically appealing method, called ``GECO'', for imposing inequality constraints while training VAEs. Here, we extend the GECO framework to deal with multiple asymmetric constraints. Concretely, we define the following posterior-derived and differentiable quantities:
\begin{align}\label{eq:q_def}
    \hat{Q}_\mathrm{density} &= \frac{1}{|\mathsf{G}_\mathrm{obj}|}\,\sum_{(l,m) \in \mathsf{G}_\mathrm{obj}} \hat{c}^{(l,m)}_\mathrm{grid},\nonumber\\
    \hat{Q}_\mathrm{area} &= \frac{1}{2|\mathsf{G}_\mathrm{nat}|}\sum_{k=1}^{\hat{K}}\hat{A}_k^\mathrm{mask} + \frac{1}{2|\mathsf{G}_\mathrm{nat}|}\sum_{k=1}^{\hat{K}}\hat{A}_k^\mathrm{BB},
\end{align}
where $\hat{A}_k^\mathrm{mask} = \sum_{(i,j) \in \mathsf{G}_\mathrm{nat}} \hat{\pi}^{(i,j)}_k$ and $\hat{A}_k^\mathrm{BB} = \hat{b}_\mathrm{w}\,\hat{b}_\mathrm{h}$ denote the mask and bounding box area of object $k$, respectively. As we will see, $\hat{Q}_\mathrm{density}$ and $\hat{Q}_\mathrm{area}$ can be used to control the number of objects and their size.
We further define $\hat{Q}_\mathrm{rec} \equiv \mathcal{L}_\mathrm{rec}$ for convenience and consider the following loss function:
\begin{multline}\label{eq:loss_sapr}
    \mathcal{L}_\mathrm{SAPR} = \mathcal{L}_\mathrm{KL} + \sum_{\beta \in \mathsf{CON}} \bigg[\overline{\lambda_\beta}\, u\left(\hat{Q}_{\beta}; Q_{\beta}^\mathrm{lo} \right) \\
    + \lambda_\beta\, \overline{v\left(\hat{Q}_{\beta}; Q_{\beta}^\mathrm{lo}, Q_{\beta}^\mathrm{hi}\right)}\bigg],
\end{multline}
where $\mathsf{CON} = \{\mathrm{rec}, \mathrm{density}, \mathrm{area}\}$, $\lambda_\beta$ are strictly positive dynamical (trainable) variables in a specified range $[\lambda_\beta^\mathrm{lo}, \lambda_\beta^\mathrm{hi}]$, $Q_\beta^{\mathrm{lo(hi)}}$ are specified lower (upper) bounds, overline implies stop-gradient operator, and:
\begin{align}\label{eq:u_v_sapr}
u(\hat{Q}; Q^\mathrm{lo}) &= \hat{Q}\, \text{sign}(\hat{Q} - Q^\mathrm{lo}) ,\nonumber\\
v(\hat{Q}; Q^\mathrm{lo}, Q^\mathrm{hi}) &= \min(\hat{Q} - Q^\mathrm{lo}, Q^\mathrm{hi} - \hat{Q}).
\end{align} 
This loss function is best understood in contrast to the usual VAE loss $\mathcal{L}_\mathrm{VAE} = \mathcal{L}_\mathrm{KL} + \mathcal{L}_\mathrm{rec}$. In $\mathcal{L}_\mathrm{SAPR}$, our primary objective is to minimize $\mathcal{L}_\mathrm{KL}$, a surrogate for the descriptive complexity of the representation, subject to posterior bounds over reconstruction error, density of foreground instances, and their area. These transparent and interpretable constraints effectively rule out undesirable modes, such as very sparse solutions or solutions in which descriptive complexity is heavily sacrificed in favor of improving the reconstruction error by an imperceptible amount. Consider the two scenarios: (1) if $\hat{Q} > Q^\mathrm{hi}$ or $\hat{Q} < Q^\mathrm{lo}$, i.e. the constraint is \textit{not} satisfied, then $v<0$ and minimization of $\mathcal{L}_\mathrm{SAPR}$ leads to an increase in the penalty strength $\lambda$ and a stronger effort toward pushing $\hat{Q}$ inside the acceptable range\footnote{Note the opposite sign of $u$ in the two cases}; (2) if $\hat{Q} \in (Q^\mathrm{lo}, Q^\mathrm{hi})$, i.e. the constraint \textit{is} satisfied, the model still tries to reduce $\hat{Q}$ but does so with less urgency: in this case $v>0$ and the penalty strength $\lambda$ tends to $\lambda^\mathrm{lo}$. The asymmetry in these equations is deliberate and reflects our preference for solutions with smallest $\hat{Q}$ within the specified bounds. Such solutions are more sparse and have lower reconstruction error.

In practice, we enforce bounds over the penalty strengths $\lambda_\beta$ by clamping them to $[\lambda_\beta^\mathrm{lo}, \lambda_\beta^\mathrm{hi}]$ after each gradient update step. We set the bounds to $\lambda_\beta^\mathrm{lo} = 0.1$ and $\lambda_\beta^\mathrm{hi} = 10$ for all three constraints. Furthermore, we set $Q^\mathrm{lo}_\mathrm{rec}\equiv 0$ so that we never degrade the reconstruction quality of the solution in case all other constraints are satisfied. The lower bounds for $\hat{Q}_\mathrm{area}$ and $\hat{Q}_\mathrm{obj}$ are chosen to be strictly positive to prevent pathologically ``empty'' solutions.\\

\noindent{\bf Warm-Up phase ---} To achieve faster training, we use a simple strategy to precondition the inference model toward recognizing regions unaccounted for by the background component as potential foreground objects. To this end, we compute $\scriptstyle\delta^{(i,j)} = \norm{\mathbf{x}^{(i,j)} - \hat{\mathbf{y}}_0^{(i,j)}}_2^2$, i.e. the pixel-wise residual between the input image and the background component. All of the proposed bounding boxes are ranked according to their average value of $\boldsymbol{\delta}$. A proposal with a high rank is associated to a image region that is poorly explained by the background and is therefore likely to correspond to a foreground object. Correspondingly, we promote its probability. 
The opposite is true for proposals of low $\delta$-rank. We implement this trick by replacing $\hat{p}_\mathrm{grid}$ obtained from the U-Net with its weighted average together with the ranks, $\hat{p}_\mathrm{grid} \leftarrow \left[1-f(t)\right]\, \hat{p}_\mathrm{grid} + f(t)\, \mathrm{rank}_\mathrm{grid} \, / \, |\mathsf{G}_\mathrm{obj}|$, where the entries of $\mathrm{rank}_\mathrm{grid}$ are integers between $1$ and $|\mathsf{G}_\mathrm{obj}|$, and $f(t)$ is kept at $0.4$ for the first several epochs and is then linearly annealed to zero.

\section{Consensus segmentation via graph community detection\label{sec:graph_segmentation}}

\begin{figure}[t]
\centering
\includegraphics[width=0.85\linewidth]{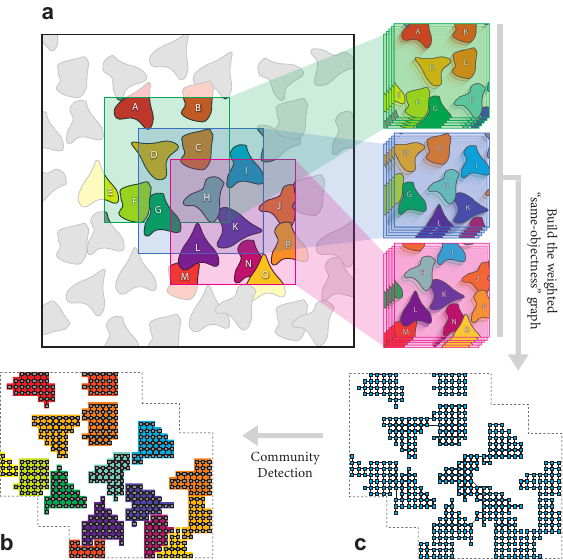}
\caption{\textbf{Consensus segmentation via graph community detection.} (a) multiple posterior samples are drawn from overlapping sliding windows; (b) the posterior samples are summarized into a weighted ``same-objectness'' graph; (c) the graph is cut using a modularity-based community detection algorithm. Please refer to Sec.~\ref{sec:graph_segmentation} for further details.}
\label{fig:graph_cutting}
\end{figure}
By construction, our approach provides a posterior probability distribution over segmentation masks. As a baseline, one may take a single point estimate $\mathbf{m}^* \, | \, \mathbf{x} = \mathrm{argmax}_{k} \hat{\boldsymbol{\pi}}_k$ obtained from a single posterior sample $\hat{\boldsymbol{\pi}}$, which usually yields a reasonable segmentation. This approach, however, disregards the valuable information encoded in the posterior distribution. A more satisfactory approach entails computing a {\em posterior averaged segmentation}, a quantity that is canonically well-defined for {\em semantic} segmentation but is ambiguous for {\em instance} segmentation. 
While several methods exists for obtaining probabilistic segmentations (e.g. see Ref.~\citemain{kohl2019probabilistic} for a probabilistic U-Net), we are not aware of any principled approach for {\em combining} posterior instance segmentation samples into a {\em consensus} segmentation. A related problem is encountered in ``stitching'' instance segmentations across sliding windows e.g. for processing large high-resolution datasets such as aerial images and cell microscopy images. The stitching problem is similar in essence as it requires a strategy for fusing segmentation across processing boundaries. Here, we introduce a method that solves both by framing it as a graph community detection problem.

Our strategy is illustrated in Fig.~\ref{fig:graph_cutting}. In brief, we run the inference on the image and obtain $N_\mathrm{post}$ posterior samples of the foreground mixing probabilities $\boldsymbol{\pi}_{1:K}$. For large images as shown schematically in Fig.~\ref{fig:graph_cutting}{\bf a}, we simply run the inference on overlapping sliding windows and obtain posterior samples from each window. Next, we build a weighted undirected graph (see Suppl. Mat. \S9 for details) in which each node represents a pixel, and pixels belonging to the same instance (in any of the posterior samples and sliding windows) have a non-zero connectivity weight. More concretely, we define $e_{p,p'} = (1/N_\mathrm{post})\sum_{j=1}^{N_\mathrm{post}}\sum_{k=1}^K \hat{\pi}^{p}_{k,j} \hat{\pi}^{p'}_{k,j}$ where $p,p' \in \mathsf{G}_\mathrm{nat}$. The weights can be stored efficiently as sparse matrices. In each posterior sampling round, pixels that belong to the background instance are ignored. Through this procedure, pixels that belong to the a same instance {\em consistently} will attain strong connections while pixels that are sometimes assigned to different instances will have weaker connections. In effect, this procedure yields a ``same-objectness'' graph. We obtain the {\em consensus} segmentation by performing a fast modularity-based community detection of $\{e_{p,p'}\}$ (see e.g. the Leiden algorithm~\citemain{leiden_main}).

\section{Experiments}\label{sec:experiments}

\begin{figure}
\centering
\includegraphics[width=\linewidth]{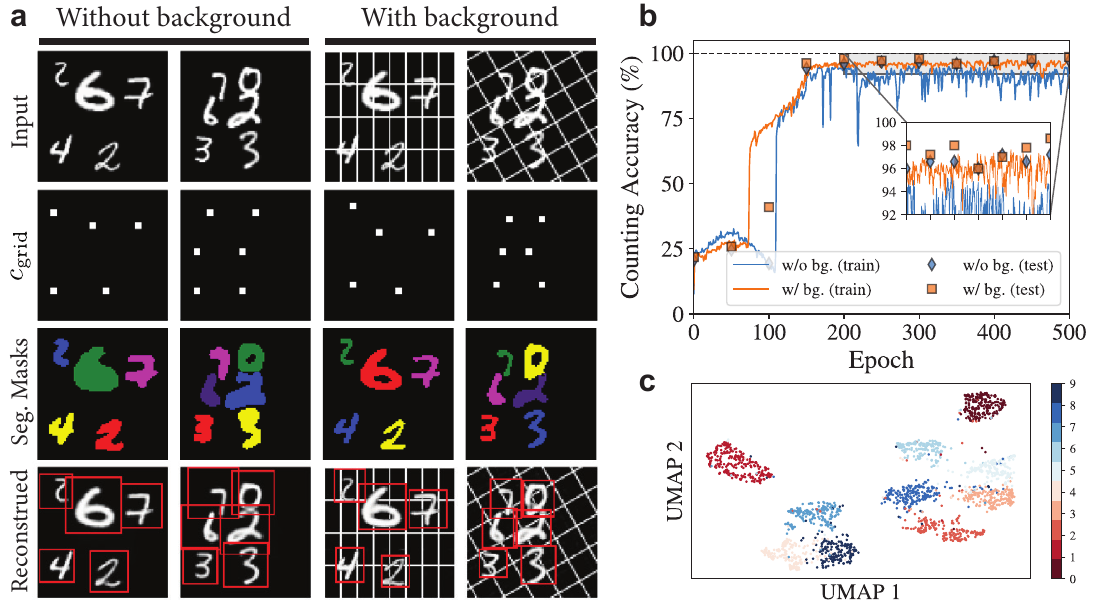}
\caption{\textbf{Model evaluation on a multi-MNIST synthetic dataset.} (a) two test input images from each dataset (w/ and w/o background) together with object detection grid, and a posterior sample of segmentation masks and reconstructed image; (b) counting accuracy vs. training epoch; (c) formation of tight clusters in the UMAP embedding of the appearance latent codes of the foreground instances.}
\label{fig:multi_mnist}
\end{figure}

We tested our framework on two multi-MNIST synthetic datasets with a black or structured background, and a cell nuclei fluorescent microscopy dataset containing $3 \times 10^4$ nuclei (``DAPI'' stains). The only pre-training strategy is the one descried as ``Warm-Up Phase'' in Sec.~\ref{sec:learning}. In all cases, we optimize $\mathcal{L}_\mathrm{SAPR}$ using the Adam optimizer~\citemain{adam_optimizer} with an initial learning rate of $10^{-3}$, and $\beta_1=0.9, \beta_2=0.999$. For the nuclei experiment, the learning rate is reduced by a factor of $0.75$ every $500$ epochs. All training was done on a single NVIDIA Tesla P100 GPU.\\

\noindent{\bf Multi-MNIST experiment ---} We generated a dataset consisting of grayscale images with size $80\mathrm{px} \times 80\mathrm{px}$  ($5000$ training, $500$ testing). Each image consists of 2 to 6 digits in equal proportions, and the digits have random sizes.  We further consider a more challenging variation of this dataset with a structured background composed of a regular grid with variable spacing and four different orientation angles chosen at random. For both datasets, we specify that the \textit{average} number of instances must be between $1.5$ and $6.5$ and that the fraction of foreground pixels must be between $5\%$ and $15\%$. We set the SAPR bounds to reflect these values. The true values are in fact $4$ and $11\%$,
respectively.
Fig.~\ref{fig:multi_mnist}\textbf{a} shows two random test cases from each dataset along with detected objects, segmentation masks, and a posterior reconstruction sample. The result is obtained after $500$ epochs of training. We observe that the counting accuracy surpasses $95\%$ in both experiments after as few as $200$ training epochs (see panel \textbf{b}). Intriguingly, the experiment with structured background yields {\em tighter} segmentation masks and consequently a slightly higher accuracy. This counter-intuitive observation can be reasoned as follows: a looser segmentation mask would require the latent space of foreground instances to encode part of the background, increasing $\mathcal{L}_\mathrm{KL}$. In the case with featureless background, there is no such incentive to learn tightly fitted segmentation masks, ultimately leading to the fallout of some legitimate proposals in case of overlapping digits due to the NMS mechanism, and ultimately a slightly lower counting accuracy. The jump in accuracy after about $100$ epochs can be traced back to the adaptive reduction of $\lambda_\mathrm{density}$ from its initial value of $1.0$ to $0.1$, resulting in more digits to be recognized. Finally, in Fig.~\ref{fig:multi_mnist}\textbf{c}, we show the UMAP embedding~\citemain{umap_original} of the instance latent codes $\mathbf{z}$ for the experiment with a structured background. The embedding is colored by the true digit labels. Formation of tight clusters is a direct result of the disentanglement of the appearance latent codes $\mathbf{z}$ from location and size latent codes $\mathbf{v}$ and the background, a property than can be helpful in downstream tasks such as instance identity clustering, while also demonstrating the excellent representation learning capability of the model. Additional details are presented in Suppl. Mat. \S10.\\

\begin{figure}
\centering
\includegraphics[width=\linewidth]{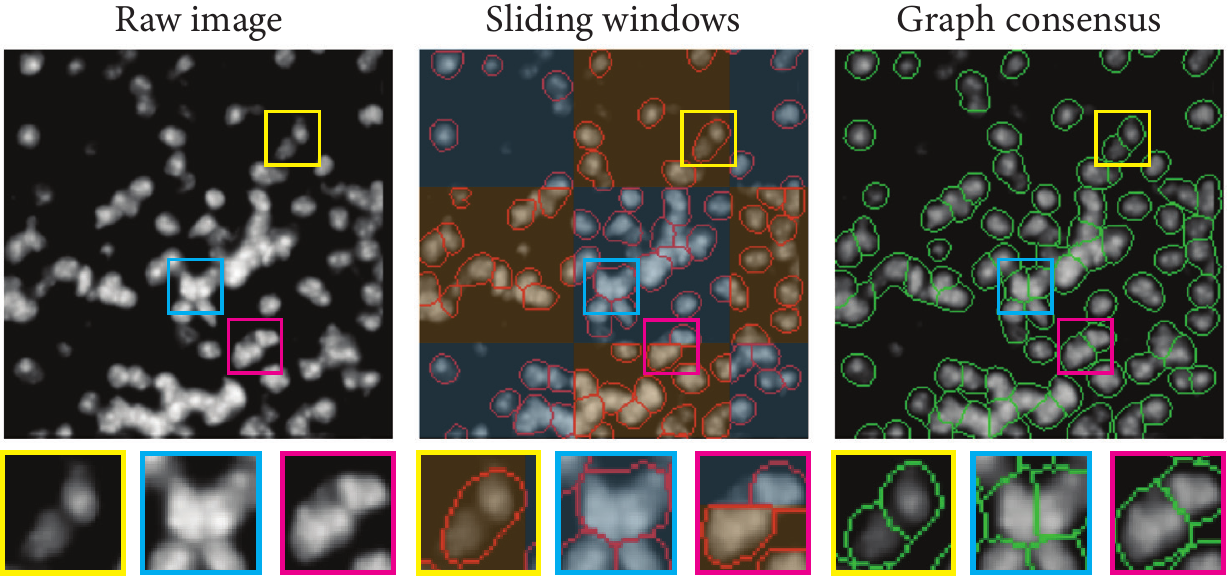}
\caption{\textbf{Segmentation of DAPI stained microscopic image.} (left) input test image [small sample]; (middle) point-estimate of segmentation obtained using non-overlapping sliding windows [shaded for clarity]; (right) consensus graph-based segmentation using overlapping sliding windows.}
\label{fig:DAPI}
\end{figure}

\noindent\textbf{Cell nuclei segmentation experiment ---} The cell nuclei dataset consists of a single high-resolution large image of linear dimension $2\times 10^5$ pixels containing approximately $3\times 10^4$ DAPI stained nuclei. We downsample the image 8-fold for faster training, and train the model with random crop mini-batches of size $80\mathrm{px}\times 80\mathrm{px}$. We estimate that between $10\%$ and $15\%$ of the pixels belong to nuclei and that the average number of nuclei in a processing window is between $5$ and $10$. We set the SAPR bounds to these crude estimates. The model is trained for 5 hours. At test time, we process the entire image using an overlapping sliding window procedure, such that every pixel is effectively processed $16$ times. In Fig.~\ref{fig:DAPI} we show a small, denser than average, portion of the input image along with the obtained segmentations. The middle panel shows a point estimate segmentation from a set of non-overlapping processing windows; note the boundary artifacts around the edges and the mistakes in the interior of the processing windows (see the zoomed regions). In the right panel, we show the segmentation obtained from overlapping sliding windows combined with the graph community detection algorithm; note that both issues have been resolved. Additional results are provided in Suppl. Mat. \S11.\\

\section{Conclusions\label{sec:conclusion}}
In conclusion, we introduced CellSegmenter, a deep generative model and a variational inference strategy with a structured latent space tailored for unsupervised representation learning and instance segmentation of modular images. CellSegmenter inference is parallel, without any recurrent units, and is able to resolve object-object occlusion while simultaneously treating distant instances independently. This leads to fast training times and a favorable scaling with the number of objects. We have demonstrated that CellSegmenter learns to count with high accuracy for both featureless and structured backgrounds and that the learned object representations are disentangled, a helpful property for downstream tasks such as classification and clustering. 

CellSegmenter is able to segment microscopic images which are challenging for human experts. It does so by leveraging three novel strategies: (1) powerful amortized inference algorithm based on U-Net, (2) posterior regularization, and (3) graph-based segmentation for leveraging posterior uncertainties. We believe these strategies will prove beneficial in a wider range of applications. In the context of segmenting cell microscopy data, we believe that the obtained segmentations can be readily improved using the current model by: (1) increasing the capacity of the networks and using full-resolution images, (2) utilizing partial annotations, and (3) using complementary data modalities (e.g. cell membrane stains, gene expression, etc) simultaneously as additional input image channels.



{\small
\bibliographystylemain{unsrt}
\bibliographymain{CellBender}
}

\setcounter{equation}{0}
\setcounter{figure}{0}
\setcounter{table}{0}
\setcounter{page}{0}
\setcounter{section}{0}
\renewcommand{\thesection}{S\arabic{section}}
\renewcommand{\theequation}{S\arabic{equation}}
\renewcommand{\thefigure}{S\arabic{figure}}

\clearpage

\begin{center}
\textbf{\Large Supplemental Materials}
\end{center}

\section{Determinantal Point Processes \label{sec:appendix_DPP}}
Determinantal Point Processes (DPPs) were first formalized as a general class of stochastic processes in 1975 by Macchi~\citesm{macchi1975coincidence} even though specific instances of DPPs had previously appeared in random matrix theory and quantum physics. There has been renewed interest in further characterizing the algebraic properties of DPPs and also applying these processes for machine learning tasks, as a means to encourage sample diversity and to avoid repetition (e.g. for text and video summarization tasks). The attractive properties of DPPs include exact sampling, polynomial-time inference and calculation of the partition function, and a large number of useful algebraic properties that allow effortless marginalization and conditioning. None of the other generic anti-correlated point processes, e.g. Mat\'ern repulsive processes and Markov random fields with negative correlations, allow polynomial-time inference and partition function calculation. Please refer to~\citesm{kulesza2012determinantal_suppl} for a recent review.


In this paper, we used DPPs to model the anti-correlation between foreground instances in our generative model. In many modular images, e.g. aerial images of buildings and cell microscopy, basic laws of physics precludes overlapping objects. As such, it is desirable for a generative model to represent this prior structure by depleting the probability mass of unphysical configurations. 

Formally, a DPP defined over a finite set $\Omega$ is specified by a positive semi-definite similarity kernel $\mathcal{S}: \Omega \times \Omega \rightarrow \mathbb{R}$, and the probability of selecting $\omega \subseteq \Omega$ is given as:
\begin{equation}\label{eq:dpp_defn}
    P_\mathrm{DPP}(\omega \subseteq \Omega) = \frac{\det(\mathcal{S}_\omega)}{\det(\mathcal{S}_\Omega + \mathbf{I}_{\Omega \times \Omega})}, 
\end{equation}
where $\mathcal{S}_\omega$ is the square sub-matrix of $\mathcal{S}_\Omega$ obtained by keeping only rows and columns indexed by $\omega$. The partition function is the denominator in Eq.~\eqref{eq:dpp_defn}. Additional details and more general definitions of DPPs can be found in Ref.~\citesm{kulesza2012determinantal_suppl}.

\begin{figure}
\centering
\includegraphics[width=\linewidth]{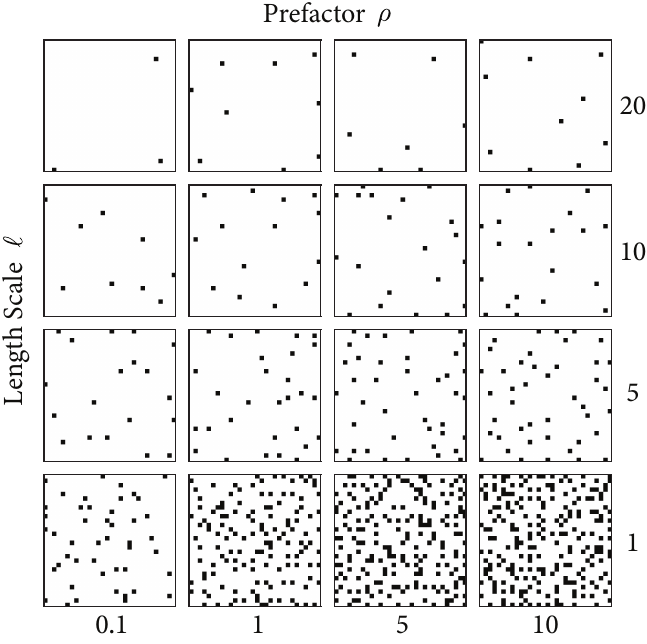}
\caption{\textbf{DPP with RBF similarity kernel.} A sample is shown for each choice of parameters $(\rho, \ell)$.}
\label{fig:dpp}
\end{figure}

In our application, $\Omega \equiv \mathsf{G}_\mathrm{obj}$ i.e. the coarse object grid and $\mathcal{S}^{l,m} = \rho\,\exp\big[-\norm{\mathbf{r}_l - \mathbf{r}_m}_2^2 / (2\,{\ell}^2)\big]$, $\mathbf{r}_l, \mathbf{r}_m \in \mathsf{G}_\mathrm{obj}$, is the RBF kernel with two learnable parameters, $\ell$ and $\rho$. Intuitively, $\rho$ and $\ell$ control the density and length scale of objects over the grid. Fig.~\ref{fig:dpp} shows samples drawn from this distribution for several choices of $\rho$ and $\ell$; note the repulsion between the chosen grid points.

Conveniently, we do not need to sample from DPP during model training or inference; rather, we only need the ability to calculate $P_\mathrm{DPP}(\omega)$ efficiently and differentiably\footnote{With respect to the parameters $\rho$ and $l$ of the DPP similarity kernel.} for any $\omega$ proposed by the inference process~\footnote{If sampling is needed, e.g. for drawing {\em de novo} images from the trained generator, it can also be done exactly in polynomial time~\citesm{kulesza2012determinantal_suppl}}. Calculating $P_\mathrm{DPP}$ involves two determinant evaluations, each of which can be done in $\mathcal{O}(|\mathcal{G}_\mathrm{obj}|^3)$ time via Cholesky decomposition\footnote{We simply use the PyTorch \texttt{logdet} function which provides support back-propagation out of the box.}. We treat both $\rho$ and $\ell$ as learnable parameters. The gradient signals can back-propagate through the Cholesky factorization and reach $\rho$ and $\ell$.

\section{Deforming $\mathbb{R}^4$ to Bounding Boxes \label{sec:appendix_BB}}
As mentioned in Sec.~\ref{sec:generative_model}, we obtain the instance bounding boxes by sampling $\mathbf{v}_{1:K} \sim \mathcal{N}(\mathbf{0}_4, \mathbf{I}_{4\times4})$ and  transforming it; the explicit transformation is given as follows:
\begin{equation}
\begin{split}
     t_k^\mathrm{x}, t_k^\mathrm{y}, t_k^\mathrm{w}, t_k^\mathrm{h} & = \mathrm{SAF}\left(\mathbf{v}_k; \boldsymbol{\theta}_\mathrm{b},\boldsymbol{\Theta}_\mathrm{w} \right),\\
      b^\mathrm{x}_k & = \frac{W}{\tilde{W}} \left(i^\mathrm{x}_k + t_k^\mathrm{x} \right), \\
      b^\mathrm{y}_k & = \frac{H}{\tilde{H}} \left(i^\mathrm{y}_k + t_k^\mathrm{y} \right), \\
      b^\mathrm{w}_k & = \ell_\mathrm{obj}^< + \left(\ell_\mathrm{obj}^>-\ell_\mathrm{obj}^<\right) t_k^\mathrm{w}, \\
       b^\mathrm{h}_k & = \ell_\mathrm{obj}^< + \left(\ell_\mathrm{obj}^>-\ell_\mathrm{obj}^<\right) t_k^\mathrm{h}, \\
       \mathbf{B}_k & = \left(b^\mathrm{x}_k, b^\mathrm{y}_k, b^\mathrm{w}_k, b^\mathrm{h}_k\right),
\end{split}
\end{equation}
where $\mathrm{SAF}(\mathbf{v}; \boldsymbol{\theta}_\mathrm{b},\boldsymbol{\Theta}_\mathrm{w}) \equiv \mathrm{sigmoid}(\boldsymbol{\theta}_\mathrm{b} + \boldsymbol{\Theta}_\mathrm{w} \mathbf{v})$ is an affine transformation followed by a sigmoid, implementing the mapping from $\mathbb{R}^4$ to $(-1, 1)^4$, $(i^\mathrm{x}_k, i^\mathrm{y}_k)$ are the discrete coordinates of the $k$'th object on the coarse grid $\mathsf{G}_\mathrm{obj}$, and $\ell_\mathrm{obj}^<$, $\ell_\mathrm{obj}^>$ are specified hyperparameters that determine the range of permissible sizes for the bounding boxes; $\boldsymbol{\theta}_\mathrm{b} \in \mathbb{R}^4$ and $\boldsymbol{\Theta}_\mathrm{w} \in \mathbb{R}^{4\times 4}$ are learnable model parameters.

\section{Architectures of the Foreground Appearance Decoder and Encoder\label{sec:appendix_fg_decoder}}
The foreground appearance decoder $\mathrm{DEC}_\mathrm{fg}$ was defined in Eq.~\eqref{eq:dec_fg} and maps $\mathbf{z} \in \mathbb{R}^{D_\mathrm{fg}}$ to: (1) a rendering of the instance appearance, $\mathbf{y}^c \in \mathsf{IM}^c \equiv \mathbb{R}^{C \times H_\mathrm{fg} \times W_\mathrm{fg}}$, and (2) mixing weights $\mathbf{w}^c \in \mathsf{W}^c \equiv (0, 1)^{H_\mathrm{fg} \times W_\mathrm{fg}}$. Since we expect a strong correlation between an object's appearance and mixing weight (``soft mask''), we expect to benefit from weight sharing.

We have chosen the following values in our experiments: $D_\mathrm{fg} = \pink{20}$, $C = \pink{1}$, $W = \pink{80}$, $H = \pink{80}$, $W_\mathrm{fg} = \pink{28}$, and $H_\mathrm{fg} = \pink{28}$. The reference implementation of $\mathrm{DEC}_\mathrm{fg}$ is as follows. First, we map $\mathbf{z}$ to a low resolution grid and progressively expand it via a sequence of transposed convolutions to an image in $C + 1$ channels; we take the first $C$ channels as $\mathbf{y}^c$; the last channel is further transformed by a sigmoid to the unit interval $(0, 1)$ and is taken as mixing weights $\mathbf{w}^c$. The concrete implementation of the layers is:
\begin{formal}
$\mathrm{DEC}_\mathrm{fg}: [20] \rightarrow [2, 28, 28]$\\
\\
$~~~~~~~~\mathrm{Linear}(20, 1600)$\\
$~~~~~~~~\mathrm{ReLU}$\\
$~~~~~~~~\mathrm{Reshape~[1600]~to~[64, 5, 5]}$\\
$~~~~~~~~\mathrm{ConvTransposed2D}(64, 32, 4, 2, 2)$\\
$~~~~~~~~\mathrm{ReLU}$\\
$~~~~~~~~\mathrm{ConvTransposed2D}(32, 32, 4, 2, 1)$\\
$~~~~~~~~\mathrm{ReLU}$\\
$~~~~~~~~\mathrm{ConvTransposed2D}(32, 2, 4, 2, 1)$
\end{formal}
\noindent The basic layers are defined as follows: $\mathrm{Linear}(i, o)$ is a dense layer with input and output channels $i$ and $o$, respectively, $\mathrm{ReLU}$ is the rectifier activation function, and $\mathrm{Conv(Transposed)2D}(i, o, k, s, p)$ is a (transposed) convolution layer with input and output channels $i$ and $o$, kernel size $k \times k$, stride $s$, and padding $p$.

The foreground appearance encoder $\mathrm{ENC}_\mathrm{fg}$ essentially performs the inverse operation, the reference implementation of which is as follows: the cropped feature map, with shape $C_\mathrm{U} \times H_\mathrm{fg} \times W_\mathrm{fg} = \pink{32 \times 28 \times 28}$ from the U-Net is processed through a series of 2D convolutions, resulting in a feature map with shape $\pink{64 \times 7 \times 7}$. The latter is flattened to $3136$ features; a linear readout to $D_\mathrm{fg} = \pink{20}$ dimensions yields $\mathbf{z}^\mu$; a second linear readout to $D_\mathrm{fg} = \pink{20}$ dimensions followed by a \texttt{Softplus} transformation yields $\mathbf{z}^\sigma$. The concrete implementation of the layers is:
\begin{formal}
$\mathrm{ENC}_\mathrm{fg}: [32, 28, 28] \rightarrow ([20], [20])$\\
\\
$~~~~~~~~\mathrm{Conv2D}(32, 32, 4, 1, 2)$\\
$~~~~~~~~\mathrm{ReLU}$\\
$~~~~~~~~\mathrm{Conv2D}(32, 32, 4, 2, 1)$\\
$~~~~~~~~\mathrm{ReLU}$\\
$~~~~~~~~\mathrm{Conv2D}(32, 64, 4, 2, 1)$\\
$~~~~~~~~\mathrm{ReLU}$\\
$~~~~~~~~\mathrm{Flatten~to~[3136]}$\\
$~~~~~~~~\mathrm{[Linear(3136, 20), Linear(3136, 20)]}$
\end{formal}

\section{Architectures of the Background Decoder and Encoder\label{sec:appendix_bg_decoder}}
The background decoder and encoder are implemented similarly to the foreground counterparts. We recall that $\mathrm{DEC}_\mathrm{bg}$ transforms a background latent code from $\mathbb{R}^{D_\mathrm{bg}}$ to $\mathbb{R}^{C \times W \times H}$. In out experiments, we chose $D_\mathrm{bg} = \pink{20}$, $C = \pink{1}$, $W = \pink{80}$, and $H = \pink{80}$. The reference implementation is:
\begin{formal}
$\mathrm{DEC}_\mathrm{bg}: [20] \rightarrow [1, 80, 80]$\\
\\
$~~~~~~~~\mathrm{Linear}(20, 800)$\\
$~~~~~~~~\mathrm{ReLU}$\\
$~~~~~~~~\mathrm{Reshape~[800]~to~[32, 5, 5]}$\\
$~~~~~~~~\mathrm{ConvTransposed2D}(32, 32, 4, 2, 1)$\\
$~~~~~~~~\mathrm{ReLU}$\\
$~~~~~~~~\mathrm{ConvTransposed2D}(32, 32, 4, 2, 1)$\\
$~~~~~~~~\mathrm{ReLU}$\\
$~~~~~~~~\mathrm{ConvTransposed2D}(32, 16, 4, 2, 1)$\\
$~~~~~~~~\mathrm{ReLU}$\\
$~~~~~~~~\mathrm{ConvTransposed2D}(16, 1, 4, 2, 1)$
\end{formal}
The background encoder takes the feature map from the bottom of the U-Net (with shape $\pink{512 \times 5 \times 5}$ in our experiments; see Fig.~\ref{fig:inference_model}) and transforms it back to the background latent space. The final layer ends with $2D_\mathrm{bg} = \pink{40}$ channels, the first half of which is taken as $\mathbf{z}^\mu_0$, and the second half is transformed to $\mathbb{R}_+$ via \texttt{Softplus} and is taken as $\mathbf{z}^\sigma_0$. The reference implementation of the layers is:
\begin{formal}
$\mathrm{ENC}_\mathrm{bg}: [512, 5, 5] \rightarrow ([20], [20])$\\
\\
$~~~~~~~~\mathrm{Conv2D}(512, 32, 1, 1, 0)$\\
$~~~~~~~~\mathrm{Concat[AdaptiveAvgPool2D(5, 5),}$\\
$~~~~~~~~~~~~~~~~~~~~~\mathrm{AdaptiveMaxPool2D(5, 5)]}$\\
$~~~~~~~~\mathrm{Conv2D}(64, 128, 1, 1, 0)$\\
$~~~~~~~~\mathrm{ReLU}$\\
$~~~~~~~~\mathrm{Conv2D}(128, 256, 3, 1, 0)$\\
$~~~~~~~~\mathrm{ReLU}$\\
$~~~~~~~~\mathrm{Conv2D}(256, 256, 3, 1, 0)$\\
$~~~~~~~~\mathrm{ReLU}$\\
$~~~~~~~~\mathrm{Conv2D}(256, 40, 1, 1, 0)$\\
$~~~~~~~~\mathrm{Split~channels~to~([20], [20])}$
\end{formal} 
\noindent where $\mathrm{AdaptiveAvgPool2D}$ and $\mathrm{AdaptiveMaxPool2D}$ are adaptive average and max pooling operations to output dimensions $(5, 5)$. For the reference U-Net architecture provided here and used in the presented experiments, the adaptive layers are immaterial as the spatial dimension of the bottom U-Net features coincides with the output dimension of the adaptive pooling layers, $(5, 5)$. The presence of these adaptive layers, however, allows us to modify the hyperparameters of the U-Net while keeping the architecture of $\mathrm{ENC}_\mathrm{bg}$ intact.

\section{Non-Max Suppression (NMS) Operation and Intersection-over-Min\label{sec:appendix_nms}}

\begin{figure}[h!]
\centering
\includegraphics[width=\linewidth]{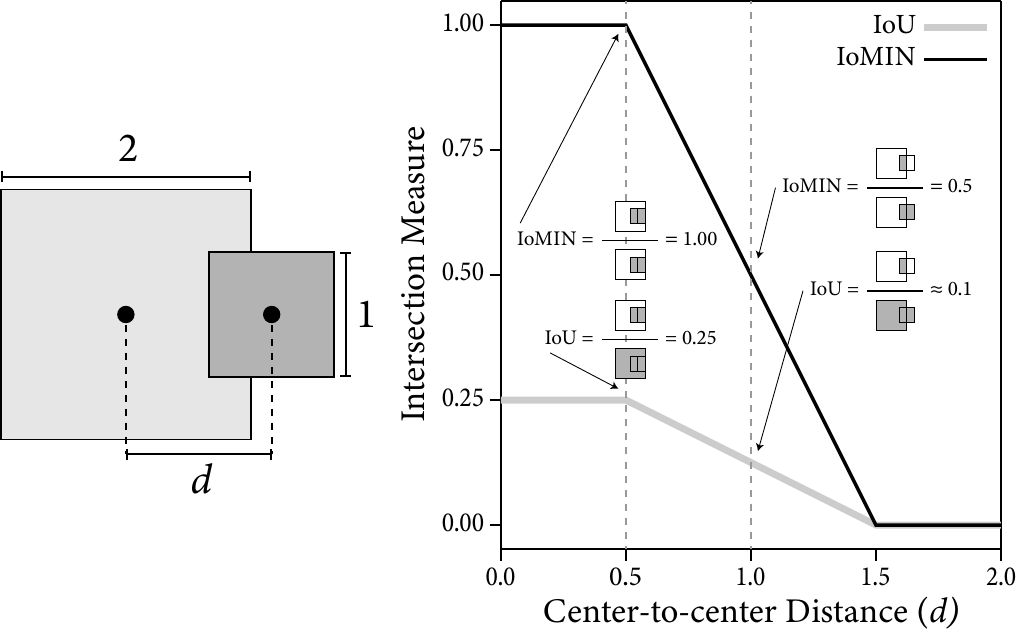}
\caption{\textbf{Intersection-over-Min (IoMIN) vs. Intersection-over-Union (IoU).} In situations where the region proposals have different sizes, IoU can produce deceptively small values, even if one proposal is entirely engulfed by the other, making this intersection measure a poor choice for threshold-based filtering of overlapping proposals. In contrast, IoMIN exhibits the desired behavior regardless of the size of the bounding boxes. Please refer to Sec.~\ref{sec:appendix_nms} for additional context.}
\label{fig:iou}
\end{figure}

In this section, we briefly provide additional details for the NMS operator mentioned in the main text, see Eq.~\eqref{eq:nms}. We recall that the NMS operator was used in the inference procedure for two reasons: (1) as a mechanism to induce negative spatial correlations between the proposals and to reflect the DPP used in the prior; (2) to remove redundant bounding box proposals from nearby grid points that may recognize the same foreground instance with high probability. 

The NMS operator takes for input (1) the provisional posterior object presence binary field $\tilde{c}_\mathrm{grid}$, (2) all possible bounding box proposals $\mathbf{B}_\mathrm{grid}$, and (3) the proposal probabilities $p_\mathrm{grid}$. First, we define a {\em score} for every point in $\mathsf{G}_\mathrm{obj}$ as follows:
\begin{equation}
    s_\mathrm{grid} = \tilde{c}_\mathrm{grid} + p_\mathrm{grid}.
\end{equation}
This scoring scheme guarantees that all of the provisionally ``on'' grid points, i.e. $\mathsf{ON} \equiv \{\mathbf{r} \in \mathsf{G}_\mathrm{obj}\, | \, \tilde{c}_\mathrm{grid}(\mathbf{r}) = 1\}$, take precedence over the provisionally ``off'' grid points, i.e. $\mathsf{OFF} = \{\mathbf{r} \in \mathsf{G}_\mathrm{obj}\, | \, \tilde{c}_\mathrm{grid}(\mathbf{r}) = 0\}$. The competition among proposals in each set, $\mathsf{ON}$ and $\mathsf{OFF}$, is resolved in favor of the one with the higher probability. If two bounding boxes overlap above a specified threshold $\alpha$, only the proposal with the highest score is allowed to pass.

The intersection measure we use here is {\em Intersection-over-Min area} (IoMIN), in contrast to the more commonly used {\em Intersection-over-Union} (IoU). These two measures are schematically compared in Fig.~\ref{fig:iou}. While the two are similar in that both quantify the overlap between bounding boxes as a scalar in $[0,1]$, $\mathrm{IoU} \le \mathrm{IoMIN}$ and the two measures behave very differently when the boxes have different sizes. The situation corresponding to one box being twice the size of the other (along each dimension) is shown in Fig.~\ref{fig:iou}. While IoU is a decent measure of agreement between a proposed bounding box and a reference bounding box (e.g. ground truth), it can be problematic for filtering overlapping proposals by thresholding: by choosing a threshold based on IoMIN, two fully encompassing boxes are guaranteed to enter a competition and only one is allowed to pass. In contrast, the IoU measure for such a configuration can be deceptively small and allow both proposals to be processed further.

At training time, we set the threshold $\alpha = 0.3$. During the course of training, we typically observe that the object presence probabilities $p_\mathrm{grid}$ tends to {\em binarized} values, with few high probability grid points corresponding to distinct instances and the vast majority of grid points having vanishing probabilities. At this point in training, the NMS operator becomes virtually the identity operator.

This auto-regulatory behavior has a simple and pleasing explanation: early into the training, the probabilities of high-quality proposals that pass the NMS filter get reinforced while the sparsity posterior regularization of $\hat{Q}_\mathrm{density}$ (see Eq.~\ref{eq:q_def}) attenuates the probability of proposals that get blocked by the NMS filter. In other words, the combination of NMS and a sparsity-inducing regularization act as a teacher mechanism and implicitly train the U-Net to avoid proposing bounding boxes that ultimately get blocked by NMS.

Even though NMS virtually deactivates itself in a fully trained model, challenging configurations involving multiple overlapping objects can still trick U-Net into producing redundant proposals. We keep NMS explicitly enabled at test time. Aiming for high detection sensitivity, we use a more permissive threshold of $\alpha=0.5$ in order to allow all instances to be detected, even if they are strongly overlapping. Posterior sampling and consensus segmentation using the graph-based strategy outlined in Sec.~\ref{sec:graph_segmentation} will effectively merge the overlapping proposals.

Finally, we recall that the number of instances $K$, derived from $c_\mathrm{grid}$, varies from image to image. Even though this variability does not pose a fundamental problem, fast GPU-based training and inference relies on efficient batching and using non-ragged tensors. To achieve batching, we choose a reasonably large upper cutoff $K_\mathrm{max}$ and always choose top-$K_\mathrm{max}$ proposals according to the scores $s_\mathrm{grid}$ among the proposals that pass the NMS filter. In most images, the actual number of proposals is lower than $K_\mathrm{max}$ and the ``extra" proposals need to be masked. We do so by multiplying the mixing weights of \textit{all} proposals by the corresponding value of $c_\mathrm{grid}$, i.e. $\mathbf{w}_j \to c_j \mathbf{w}_j$ for $j=1,\dots,K_\mathrm{max}$. 
Note that $c_j=0$ for the ``extra" proposals and therefore, this procedure masks them out so long as image reconstruction is concerned. However, gradients can still back-propagate through $c$ and reach $\mathbf{p}_\mathrm{grid}$, so that proposals which are useful to reconstruct the image are reinforced and proposals which are not helpful are further suppressed.



\section{Parallel Inference and Learning Interaction-aware Feature Maps\label{sec:appendix_fmap}}
\begin{figure}
\centering
\includegraphics[width=0.8\linewidth]{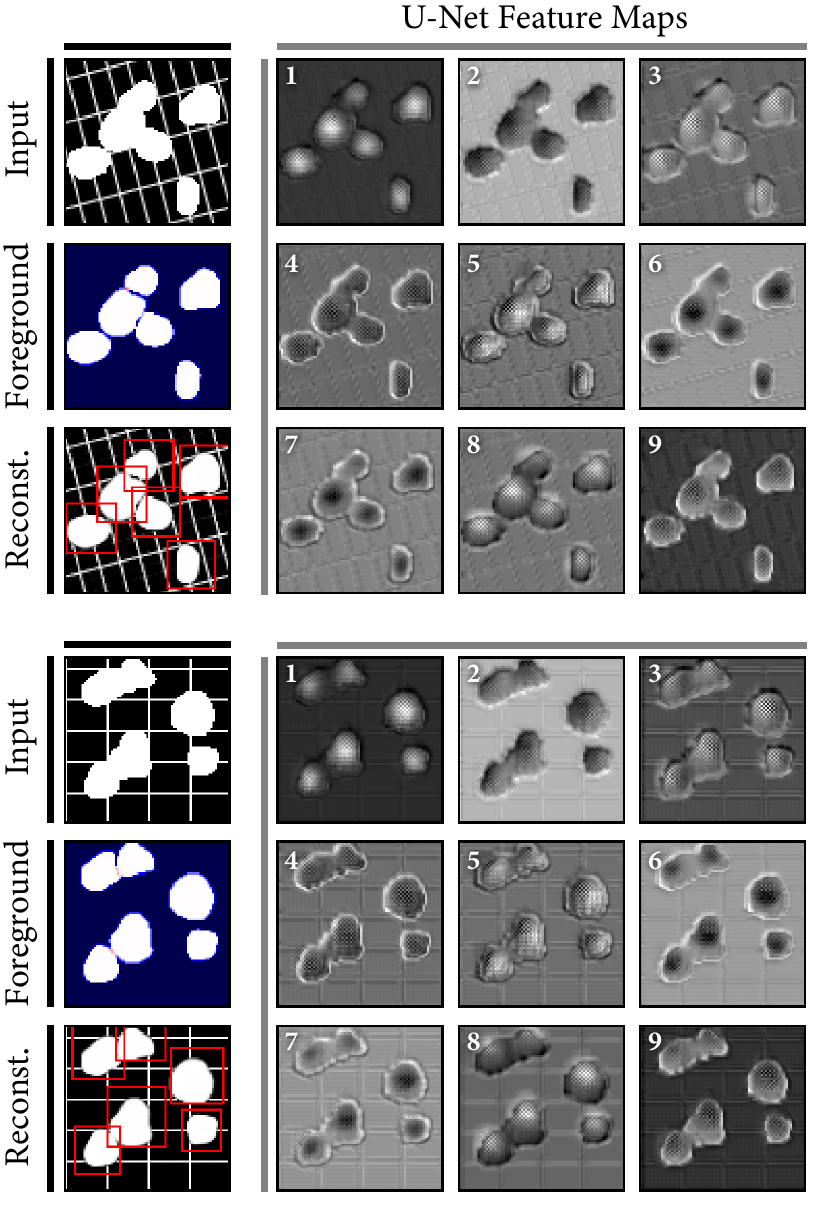}
\caption{\textbf{Opening the blackbox of amortized variational inference.} U-Net learns interaction-aware feature maps from featureless objects that provide no edge hints. Please refer to Sec.~\ref{sec:appendix_fmap} for details.}
\label{fig:fmap}
\end{figure}

The U-Net CNN architecture is a popular and powerful technique for supervised semantic segmentation tasks~\citesm{unet_original_suppl}. By combining global features and local high-resolution features, U-Net generates state-of-art semantic segmentations with simple end-to-end training. As outlined in the main text, we utilize U-Net in building a parallelized amortized variational inference framework. Furthermore, we claimed that our convolutional inference framework exhibits the highly desirable property of iterative (recurrent) inference strategies: the ability to learn object-object interactions. 

Here, we perform a simple experiment to demonstrate that the U-Net architecture indeed learns high-resolution \textit{interaction-aware} feature maps in our model. We consider a synthetic dataset similar to the multi-MNIST dataset presented in the main text but, however, with ``blobs'' instead of digits\footnote{The replacement of digits with blobs makes it easier to create a dataset with a high degree of occlusions.}. The blobs have constant intensity set to 1 and are generated by random parameterization of a finite Fourier series representing the radial distance of the boundary in polar coordinates from a randomly chosen origin. We augment the SAPR loss function, $\mathcal{L}_\mathrm{SAPR}$ (see Eq.~\ref{eq:loss_sapr}), with an object overlap penalty term:
\begin{equation}
    \mathcal{L}_\mathrm{overlap} \equiv \lambda_\text{overlap} \sum_{p \in \mathsf{G}_\mathrm{nat}} O^p, \qquad O^p \equiv \sum_{\substack{k,k'=1 \\ k \neq k'}}^K w^p_k w^p_{k'}
\end{equation}
where $\mathbf{w}_k$ are the local mixing weights introduced in Eq.~\eqref{eq:mixing_weights} and $O^p$
is the pairwise overlap between all objects at pixel $p \in \mathsf{G}_\mathrm{nat}$. We observe that during training, the value of $\mathcal{L}_\mathrm{overlap}$ decreases gradually, signalling that the model is potentially learning to account for object-object interactions and producing mutually exclusive mixing weights. We recall that in our inference framework, all instances are processed in parallel by cropping the feature map produced by the U-Net using the instance bounding boxes. Therefore, we hypothesize that the feature map itself must contain the signature of object-object interactions; we directly inspect the feature maps to scrutinize this. Fig.~\ref{fig:fmap} shows $9$ of the $32$ U-Net output channels. We point out the following observations: (1) the feature maps contain edge-like features separating nearby instances, (2) the feature maps exhibit a {\em depth effect}, resembling the deep watershed transform~\citesm{bai2017deep}, and (3) the background is almost completely removed in some of the feature maps, e.g. see feature map $\textbf{1}$), implying that the U-Net has learned to subtract the background. In the ``foreground'' panel of Fig.~\ref{fig:fmap}, we show the sum of the mixing weights $\sum_k \mathbf{w}_k$. The blue, white and red colors correspond to a value of zero, one and two respectively. We observe that the overlap between the instances (red dots) is almost completely absent, and that instances which were merged in the input exhibit a single pixel separation among them.

Intriguingly, all blobs in this dataset, both in isolation and in an overlapping configuration, have constant intensity set to 1. Furthermore, the blobs have sharp edges and are devoid of any edge hints, see the ``Input'' images in Fig.~\ref{fig:fmap}. Therefore, there is no local feature which can be used to identify the contact region between the instances and the U-Net must {\em necessarily} rely on non-local information to decide how to split overlapping instances. As an example, the model can use the location of centers and the instance sizes to estimate where the likely boundary between two instances could be. Such non-local information is only available at higher depth in the U-Net, i.e. the resolution at which $p_\text{grid}$ and the bounding boxes are inferred.


Another intriguing finding is that, in this experiment, the counting accuracy increases with the addition of the overlap penalty term, suggesting that it might be beneficial include this posterior regularization as a default component in the model.

\section{Glossary of latent variables, learnable parameters, and hyperparameters\label{sec:appendix_glossary}}
%
%
%
%
%
We provide a complete glossary of the CellSegmenter parameters here for reference. These include latent variables $\mathbf{Z}$, learnable model parameters $\mathbf{\theta}$, learnable inference parameters $\mathbf{\phi}$, and hyperparameters. We recall that the distinction between latent variables and learnable parameters is that the latent variables are given a full Bayesian treatment whereas we only aim for a point estimate for learnable variables (e.g. neural network weights). For hyperparameters that are held fixed across all experiments shown in this paper, we indicate our reference choice in \pink{purple}:\\ 

\noindent {\bf Model ---}
\begin{quote}\noindent {\em Hyperparameters}: image dimensions $C \times H \times W$ $\pink{(1 \times 80 \times 80)}$; foreground raster dimensions $C \times H_\mathrm{fg} \times W_\mathrm{fg}$ $\pink{(1 \times 28 \times 28)}$; lower and upper bounds on the linear dimension of objects, $\ell_\mathrm{obj}^<$ and $\ell_\mathrm{obj}^>$; foreground appearance latent dimensions $D_\mathrm{fg}$; image reconstruction error scale $\sigma$ (see below); background image latent dimensions $D_\mathrm{bg}$; architectures of   $\mathrm{DEC}_\mathrm{fg}$ and $\mathrm{DEC}_\mathrm{bg}$ (see Sec.~\ref{sec:appendix_fg_decoder} and Sec.~\ref{sec:appendix_bg_decoder}).\\\\
\noindent {\em Learnable parameters}: the parameters of the DPP RBF kernel, $\rho$ and $\ell$; layer weights of $\mathrm{DEC}_\mathrm{fg}$ and $\mathrm{DEC}_\mathrm{bg}$.\\\\
\noindent {\em Latent variables}: object presence binary random field $c_\mathrm{grid}$; background latent code $\mathbf{z}_0$; foreground appearance latent codes $\mathbf{z}_{1:K}$; foreground bounding box latent codes $\mathbf{v}_{1:K}$; discrete segmentation mask $\mathbf{m}$; 
\end{quote}

\noindent {\bf Inference ---}
\begin{quote}\noindent {\em Hyperparameters}: U-Net depth $D_\mathrm{U}$ $\pink{(4)}$ and number of channels $C_\mathrm{U}$ $\pink{(32)}$; architectures of $\mathrm{ENC}_\mathrm{fg}$ and $\mathrm{ENC}_\mathrm{bg}$ (see Sec.~\ref{sec:appendix_fg_decoder} and Sec.~\ref{sec:appendix_bg_decoder}).\\\\
\noindent {\em Learnable parameters}: layer weights of U-Net, $\mathrm{ENC}_\mathrm{fg}$, and $\mathrm{ENC}_\mathrm{bg}$.\\\\
\noindent {\em Latent variables}: same as above.\\
\end{quote}

\noindent {\bf Learning ---}
\begin{quote}\noindent {\em Hyperparameters}: Adam optimizer $\alpha, \beta_1, \beta_2$ $\pink{(10^{-3}, 0.9, 0.999)}$; SAPR lower and upper bounds $Q^\mathrm{lo}$ and $Q^\mathrm{hi}$ for each constraint; SAPR penalty lower and upper bounds $\lambda^\mathrm{lo}$ $\pink{(0.1)}$ and $\lambda^\mathrm{hi}$ $\pink{(10)}$ for each constraint.
\end{quote}

The image reconstruction error scale $\sigma$, which is a model hyperparameter, must be thought of as a normalization factor necessary to make $\mathcal{L}_\mathrm{rec} \sim \mathcal{O}(1)$. Conveniently, the precise value of $\sigma$ is immaterial within the SAPR framework since ultimately the combination $\lambda_\mathrm{rec}/\sigma^2$ controls the magnitude of the reconstruction term, see Eqs.~\eqref{eq:loss_explicit} and~\eqref{eq:loss_sapr}. However, a rough estimate of $\sigma$ is necessary in order to avoid a poorly conditioned starting point for learning. This rough estimate can be obtained, for instance, by fitting a two-component Gaussian to the intensity histogram of raw images, identifying the foreground component, and taking its standard deviation as an upper bound for $\sigma$. Depending on the structure of the images, the intensity histogram of the foreground component may also be approximately determined via Otsu's method.

\section{Monte-Carlo Estimation of $f_\mathrm{KL}^\mathrm{grid}$\label{sec:appendix_kl_cgrid}}
As outlined in Sec.~\ref{sec:learning}, the $\mathcal{L}_\mathrm{KL}$ term in the loss function comprises the KL divergence between all posterior and prior pairs. Most of the latent variables appearing in our model have a standard Normal prior distribution and a Gaussian posterior, allowing analytical calculation of the KL divergence terms. An exception is $c_\mathrm{grid}$ which has a DPP prior and a categorical (multinomial) posterior. In this section, we provide an explicit expression to serve as an unbiased estimator of $f_\mathrm{KL}^\mathrm{grid}$, the KL divergence term associated with $c_\mathrm{grid}$. Following the definitions, we obtain:
\begin{widetext}
\label{eq:KL_cgrid}
\begin{align}
f_\mathrm{KL}^\mathrm{grid} &= D_\mathrm{KL}\left[\mathrm{Cat}(\hat{p}_\mathrm{grid})\, ||\, P_\mathrm{DPP}(\mathcal{S})\right]\nonumber\\
&= -\sum_{\omega \in 2^{\mathsf{G}_\mathrm{obj}}}P_\mathrm{Cat}(\omega \, | \, p_\mathrm{grid})\,\log \frac{P_\mathrm{DDP}(\omega \, | \, \mathcal{S})}{P_\mathrm{Cat}(\omega \, | \, p_\mathrm{grid})}\nonumber\\
&\simeq \frac{1}{n_\mathrm{MC}} \sum_{i=1}^{n_\mathrm{MC}}\sum_{(l,m) \in \mathsf{G}_\mathrm{obj}}\left[\omega_i^{(l,m)}\,\log\,p_\mathrm{grid}^{(l,m)} + \left(1 - \omega_i^{(l,m)}\right)\,\log\left(1-p_\mathrm{grid}^{(l,m)}\right)\right]\nonumber\\
&\qquad - \frac{1}{n_\mathrm{MC}} \sum_{i=1}^{n_\mathrm{MC}}\log\det\left(\mathcal{S}_{\omega_i}\right) + \log\det\left(\mathcal{S}_{\mathsf{G}_\mathrm{obj}} + \mathbf{I_{|\mathsf{G}_\mathrm{obj} \times |\mathsf{G}_\mathrm{obj}}|}\right).
\end{align}
\end{widetext}
Note that we have replaced the summation over all binary fields to a finite MC estimator on the third line, $\omega_{1:n_\mathrm{MC}} \sim \mathrm{Cat}(p_\mathrm{grid})$ are i.i.d binary fields over $\mathsf{G}_\mathrm{obj}$, and we have used Eq.~\eqref{eq:dpp_defn}. We found $n_\mathrm{MC} = 1$ to work well in practice.

\section{Instance Connectivity Graph Construction and Community Detection: {\em Implementation Details}}\label{sec:appendix_graph}

\begin{figure}
\centering
\includegraphics[width=\linewidth]{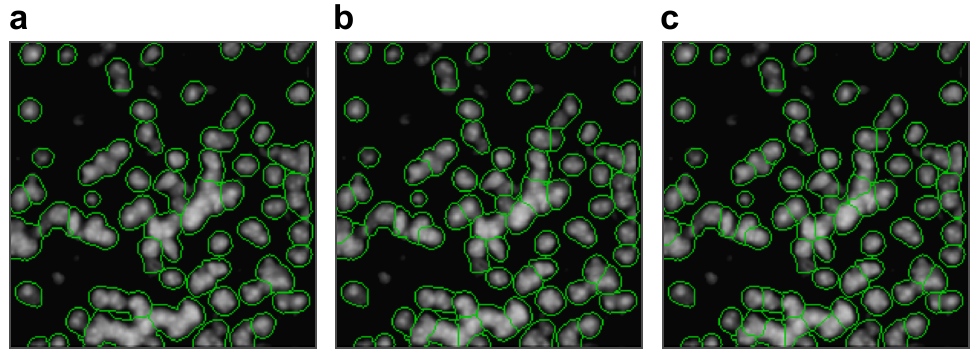}
\caption{\textbf{The effect of the modularity resolution parameter $\gamma$ on consensus segmentations.} (a) $\gamma = 100$; (b) $\gamma = 500$; (c) $\gamma = 1000$. Please refer to Sec.~\ref{sec:appendix_graph} for details.}
\label{fig:seg_res}
\end{figure}

In this section, we provide the implementation details for the graph-based consensus segmentation algorithm outlined in Sec.~\ref{sec:graph_segmentation}.

Let $\mathbf{x} \in \mathbb{R}^{C \times H_\mathbf{x} \times W_\mathbf{x}}$ be a large contiguous image that we wish to segment. In practice, $\mathbf{x}$ can be several orders of magnitude larger than the small processing window of CellSegmenter, $80\,\mathrm{px} \times 80\,\mathrm{px}$ in our reference implementation. As a first step, we generate an {\em global index matrix} $\mathbf{I}$ with the same spatial dimensions as $\mathbf{x}$, and with entries numbered $0,1,\dots,W_\mathbf{x} \times H_\mathbf{x}-1$. Next, we pad both $\mathbf{x}$ and $\mathbf{I}$: we use reflection padding for the image matrix $\mathbf{x}$, and constant padding with value $-1$ for the index matrix $\mathbf{I}$. At this stage, we process the entire image, in parallel, and by cropping overlapping sliding windows. In our reference implementation, the processing window is $80\,\mathrm{px} \times 80\,\mathrm{px}$; each window is displaced in either direction by $20~\mathrm{px}$ with respect the previous window, such that every pixel participates in $16$ inferences. For each window, we use the mixing probabilities $\boldsymbol{\pi}_{1:K}$ and the appropriate patch of the global index matrix to compute the graph weights, $e_{p,p'}=\sum_k \pi_k^p \pi_k^{p'}$ and store them as a COO sparse matrix. An efficient GPU-based implementation of this step is described below.

Since the pixel indexing is consistent across all the processing windows, the graph weights matrices for all processing windows can be simply summed together to obtain the global connectivity graph. In practice, this parallel and efficient map-reduce framework allows linear speedup with the number of available GPUs. For the example shown in Fig.~\ref{fig:dapi_global}, the graph contains $\sim 10^6$ vertices and $\sim 10^8$ edges, and takes only a few minutes to compute using a single NVIDIA Tesla P100 GPU.

Equipped with the connectivity graph, the final step is obtaining the consensus segmentation by detecting the graph communities. The communities are discovered by maximizing a properly defined metric. We have experimented using both the CPM~\citesm{CPM_community_detection} and the RB~\citesm{RBC_community_detection, RBC_community_detection_2} metrics and obtained similar results using the excellent implementation of the Leiden algorithm~\citesm{leiden_suppl,leiden_software} which provides both metrics.

Either metric admits a {\em resolution parameter} which can be roughly understood as a connectivity threshold below which a community is divided into further sub-communities, i.e. higher resolution leads to more communities. In the cell nuclei experiment, we have observed that most cells are segmented consistently for a wide range of values of the resolution parameter. However, few ambiguous regions remain sensitive to the choice of the resolution parameter, see Fig.~\ref{fig:seg_res}. We remark that the ability to control under- and over- segmentation, in a post-processing steps, is an extremely useful feature in practice. The resolution parameter provides a simple and intuitive ``knob'' to tune the level of segmentation \textit{without} the need to perform another round of time consuming and expensive model training.

We have implemented two modes in CellSegmenter for choosing the segmentation resolution parameter, an automated mode and an interactive mode. In the automated mode, a recommended value of the resolution parameter is determined by approximately maximizing the mutual information between the graph-based consensus segmentation and individual posterior samples. In the interactive mode, the user selects a small region in the image and experiments with different resolution parameters in order to determine a resolution that produces the desired segmentation. The chosen value is then used to perform community detection on the entire graph.

Finally, we emphasize that in application to segmenting cell microscopy images, having the ability to tune the segmentation stringency, as a post-processing step, is a highly desirable aspect of our method and can help ameliorate the issue of under- and over- segmentation of challenging cell microscopy images~\citesm{caicedo2019evaluation}.\\

\noindent {\bf Parallel computation of sparse connectivity weights ---} We briefly describe a fast GPU-based algorithm for calculating $e_{p,p'} = \sum_k \pi_k^p \pi_k^{p'}$ in a given processing window. Appealing to the locality of foreground instances, only nearby pixels will have a non-zero connectivity weight. Therefore, we introduce a cutoff distance $d_c$\footnote{This distance can be chosen to be a fraction of typical object size. We have found the communities to be very robust to the choice of the cutoff in practice.} and constraint the weight calculations to pixel pairs within the cutoff radius. We pad $\boldsymbol{\pi}$ and cropped index for the current processing window $\mathbf{I}^c$ with $0$ and $-1$, respectively, along each spatial dimension by $d_c$. We compute the weight between all pixels and their respective neighbors a distance $\mathbf{d} = (\delta_x, \delta_y)$ apart as follows. First, we perform a circular shift of $\boldsymbol{\pi}$ and $\mathbf{I}^c$ with displacement $\mathbf{d}$ and compute the pixel-wise dot product between the shifted and reference $\boldsymbol{\pi}$:
\begin{equation}
\mathbf{E}_\mathbf{d} \equiv \sum_{k=1}^K \mathrm{CircShift}\left(\boldsymbol{\pi}_k; \mathbf{d} \right) \odot \boldsymbol{\pi}_k
\end{equation}
where $\mathrm{CircShift}(\cdot; \mathbf{d})$ is the circular shift operator by $\mathbf{d}$ pixels. We preclude pixels such that $\mathbf{E}_\mathbf{d} < E_\mathrm{min}$, $\mathbf{I}^c = -1$, and $\mathrm{CircShift}\left(\mathbf{I}^c; \mathbf{d} \right) = -1$, and use the remaining pixels to build a COO sparse matrix. Here, we set $E_\mathrm{min} = \pink{0.01}$ to remove weak edges and reduce memory consumption. The row indices, column indices and values are read off from $\mathbf{I}^c$, $\mathrm{CircShift}\left(\mathbf{I}^c; \mathbf{d} \right)$, and $\mathbf{E}_\mathbf{d}$, respectively. This operation is repeated for half~\footnote{By symmetry argument, only one displacement for each pair $\mathbf{d}, -\mathbf{d}$ need to be considered.} of the displacement vectors with $\norm{\mathbf{d}} < d_c$, and the resulting COO sparse matrices are summed. This map-reduce operation can be efficiently implemented on GPU and, with a bit of work, can be performed in parallel for an entire processing minibatch composed of different image windows.

\section{Multi-MNIST experiment: {\em additional details}}\label{sec:appendix_mnist_additional_results}

\begin{figure}
\centering
\includegraphics[width=\linewidth]{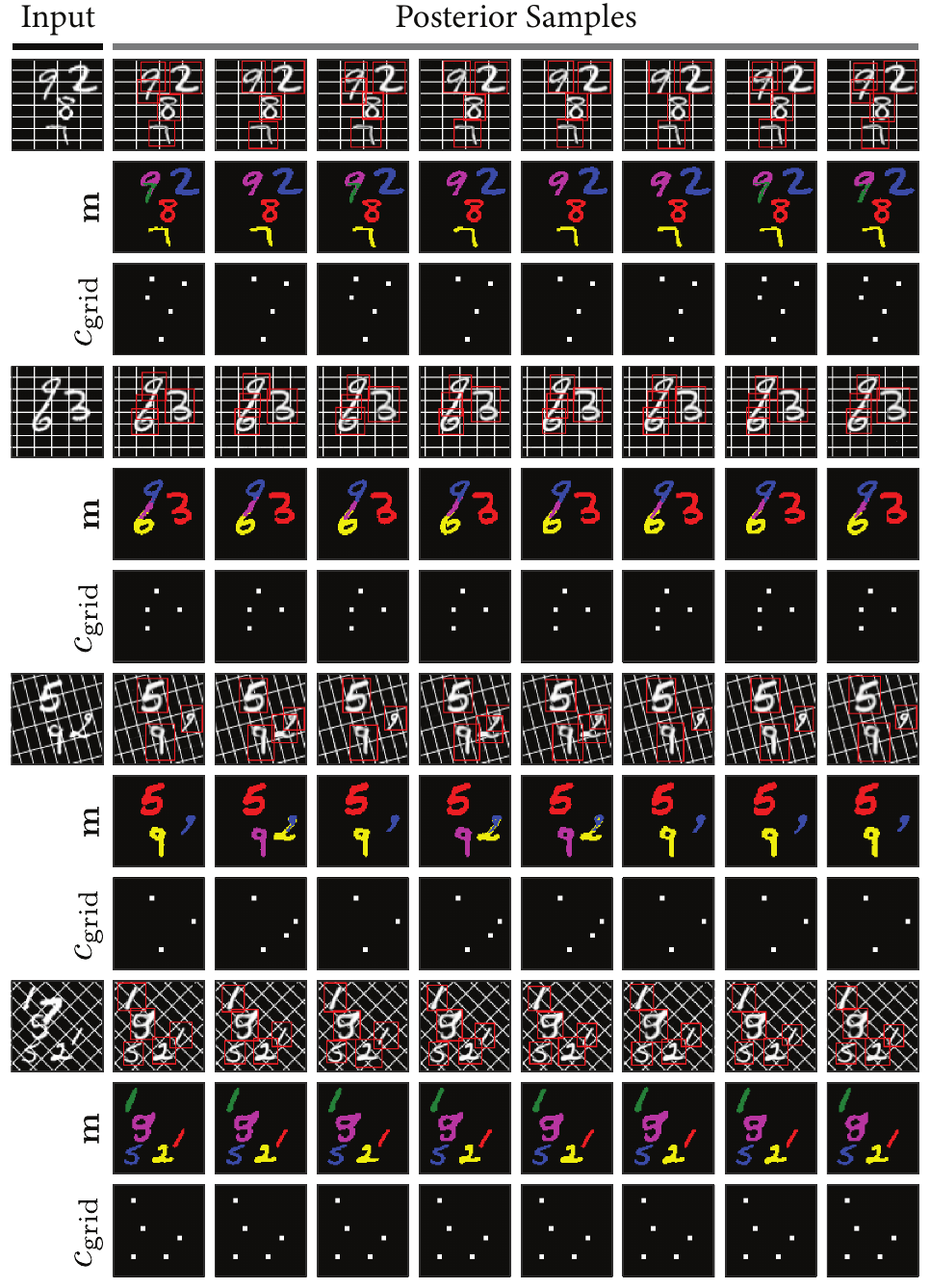}
\caption{\textbf{Additional multi-MNIST test cases.} The figure demonstrates four challenging test cases. The input image is shown in the upper left corner of each series, along with 8 posterior samples (reconstruction, segmentation mask, and object presence map). Please refer to Sec.~\ref{sec:appendix_mnist_additional_results} for details.}
\label{fig:mnist_extra}
\end{figure}

We provide supplemental details regarding the multi-MNIST experiment in this section. We train CellSegmenter according to the procedure outlined in Sec.~\ref{sec:learning}. The only pre-training strategy is the one descried as ``Warm-Up Phase''. The weights of all neural layers are randomly initialized via the Glorot scheme (also known as Xavier). Each training minibatch consists of 128 grayscale images with dimensions $80~\mathrm{px} \times 80~\mathrm{px}$. We set the image reconstruction error scale to $\sigma = 0.05$ to roughly satisfy $\mathcal{L}_\mathrm{rec} \sim \mathcal{O}(1)$. The SAPR bounds for $Q_\mathrm{area}$ and $Q_\mathrm{density}$ are chosen according to the crude estimates given in the main text, and we further set $Q^\mathrm{lo}_\mathrm{rec} = 0$ and $Q^\mathrm{hi}_\mathrm{rec} = 1$. Finally, we choose $K_\mathrm{max} = 10$. All other hyperparameters are set to the reference values specified in Sec.~\ref{sec:appendix_fg_decoder}, \ref{sec:appendix_bg_decoder}, and \ref{sec:appendix_glossary}.

Typical test cases where shown earlier in the main text for both featureless and structured background, see Fig.~\ref{fig:multi_mnist}. Here, we focus on on the multi-MNIST dataset with structured background and study four challenging test cases to better understand the confusion modes of CellSegmenter. The results are shown in Fig.~\ref{fig:mnist_extra} where for each input image, we have provided 8 posterior reconstructions, segmentation masks $\mathbf{m}$, and object presence fields $c_\mathrm{grid}$. In the first test case, we notice over-segmentation of 9 into 0 and 1 in some of the posterior samples. A similar mistake is noticed in the second test case, resulting in the detection of an extra digit 1 bridging 9 and 6. The error mode in the third test case is curious: the curly digit 2 has lead to calling an extra digit 9 in some of the posterior samples. Finally, the overlap between 5 and 7 in the last case has led to reconstructing 5 as 9 and subsequent disappearance of 7. We note that these are all among common error modes of digit classification.

Even though the accuracy of the present model is rather high ($\sim 98\%$ counting accuracy, see Fig.~\ref{fig:multi_mnist}), we did not embark on extensive hyperparameter optimization or fine-tuning in the shown experiments. We hypothesize that the accuracy can be further improved simply with longer training and using larger neural networks (encoders, decoders, and the U-Net).

\section{Cell nuclei segmentation experiment: {\em additional details}}\label{sec:appendix_dapi_additional_results}



\begin{figure}
\centering
\includegraphics[width=\linewidth]{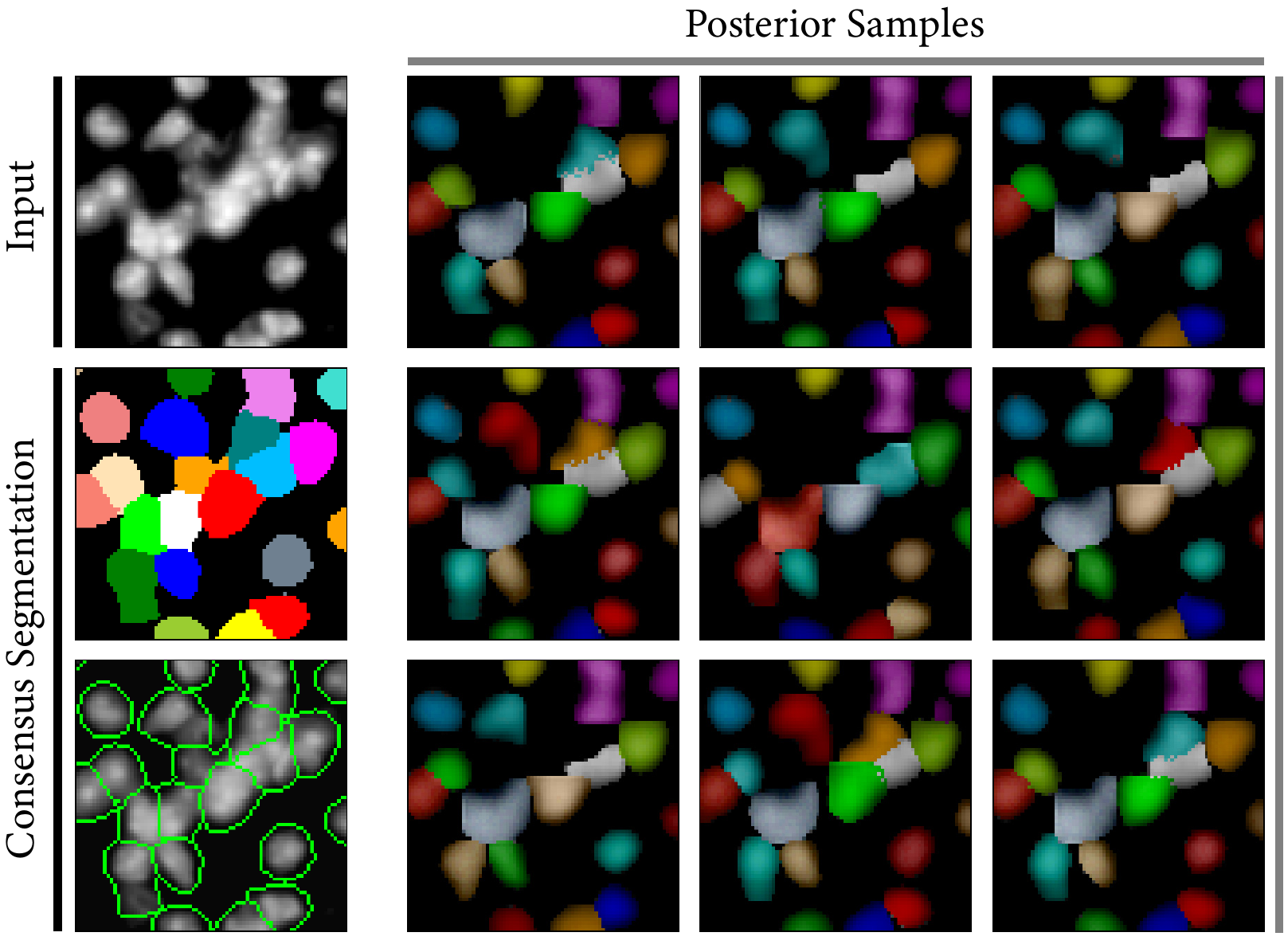}
\caption{\textbf{Cell nuclei segmentation.} The figure shows a sample input region, graph consensus segmentation, and several posterior samples (color-cycled segmentation masks superimposed on reconstructions). Please refer to Sec.~\ref{sec:appendix_dapi_additional_results} for details.}
\label{fig:dapi_extra}
\end{figure}

\begin{figure*}
\centering
\includegraphics[width=0.8\textwidth]{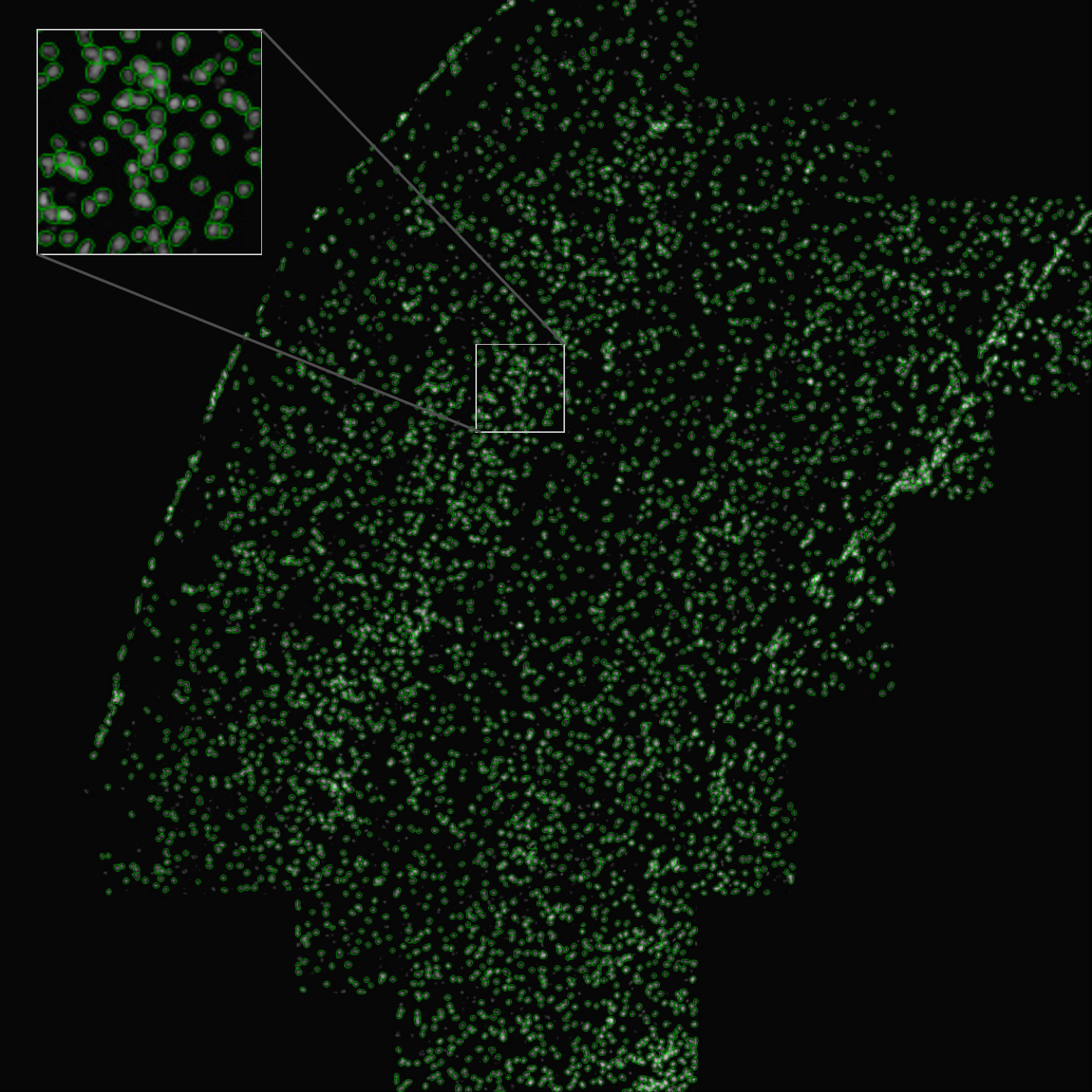}
\caption{\textbf{Global consensus segmentation of a cell nuclei dataset (DAPI stain) using CellSegmenter.} CellSegmenter can handle realistic use cases involving large contiguous images comprising thousands of instances without supervision.}
\label{fig:dapi_global}
\end{figure*}

This section includes additional details and results for the cell nuclei (DAPI stain) segmentation experiment. The dataset is publicly available from~\citesm{dapi_allen}. The major hyperparameters were given in the main text, see Sec.~\ref{sec:experiments}. The other hyperparameters were chosen similarly to the multi-MNIST experiment, see Sec.~\ref{sec:appendix_mnist_additional_results}, except for $K_\mathrm{max}$ which we set to $25$.

Fig.~\ref{fig:dapi_extra} shows the same test region as in the main Fig.~\ref{fig:DAPI} along with the consensus segmentation and nine posterior samples. The beneficial role of posterior sampling and consensus calling is noticeable: each of the posterior samples includes at least one poor decision (i.e. mergers and missed regions). The consensus segmentation, as obtained by combining a large number of posterior samples and community detection shows a remarkable improvement over individual samples, see Sec.~\ref{sec:graph_segmentation}.

Finally, Fig.~\ref{fig:dapi_global} shows the global consensus segmentation obtained for the entire dataset~\citesm{dapi_allen}, demonstrating the ability of CellSegmenter in handling realistic use cases involving large contiguous images comprising thousands of instances.

{\small
\bibliographystylesm{unsrt}
\bibliographysm{CellBender}
}

\end{document}